\documentclass{article} % For LaTeX2e
\usepackage[T1]{fontenc}
\usepackage{iclr2027_conference,times}

\usepackage{amsmath,amsfonts,bm}

\def\eqref#1{equation~\ref{#1}}
\def\1{\bm{1}}

\DeclareMathAlphabet{\mathsfit}{\encodingdefault}{\sfdefault}{m}{sl}
\SetMathAlphabet{\mathsfit}{bold}{\encodingdefault}{\sfdefault}{bx}{n}

\usepackage{hyperref}
\usepackage{url}
\usepackage{microtype}
\usepackage{graphicx}
\usepackage{adjustbox}
\usepackage{subcaption}
\usepackage{booktabs} % for professional tables
\usepackage{colortbl}
\usepackage{xcolor}
\definecolor{tablehighlight}{RGB}{186, 186, 233}
\usepackage{multirow}
\usepackage{threeparttable}
\usepackage{amssymb}
\usepackage{algorithm}
\usepackage{algorithmic}
\usepackage{wrapfig}

\renewcommand{\footnoterule}{%
  \kern -3pt
  \hrule width 0.25\columnwidth
  \kern 2.6pt
}

\title{SpectralCache: Accelerating Diffusion-Based World Models via Spectral Feature Caching}

\author{
{\fontsize{8.5}{10}\selectfont\textbf{Zhendong Mi$^{1}$, Pu Zhao$^{2}$, Ziyu Hu$^{1}$, Xiaodong Yu$^{1}$, Yanzhi Wang$^{2}$, Grace Li Zhang$^{3}$, Shaoyi Huang$^{1}$\thanks{\normalfont Corresponding author.}}}\\[0.35em]
\normalfont\small
$^{1}$Stevens Institute of Technology \quad
$^{2}$Northeastern University \quad
$^{3}$Technische Universit\"at Darmstadt\\[0.25em]
\texttt{\{zmi2,zhu31,xyu38,shuang59\}@stevens.edu}\\
\texttt{\{p.zhao,yanzhiwang\}@northeastern.edu}\\
\texttt{grace.zhang@tu-darmstadt.de}
}

\iclrfinalcopy
\begin{document}

\maketitle

% Use the paper title as the running header and keep boxed hyperlinks in the
% same standard sky blue used by the companion arXiv version.
\lhead{\scriptsize SpectralCache: Accelerating Diffusion-Based World Models via Spectral Feature Caching}
\hypersetup{
    colorlinks=false,
    pdfborder={0 0 1},
    citebordercolor={0.5294 0.8078 0.9216},
    linkbordercolor={0.5294 0.8078 0.9216},
    urlbordercolor={0.5294 0.8078 0.9216},
    pdftitle={SpectralCache: Accelerating Diffusion-Based World Models via Spectral Feature Caching},
    pdfauthor={Zhendong Mi, Pu Zhao, Ziyu Hu, Xiaodong Yu, Yanzhi Wang, Grace Li Zhang, Shaoyi Huang}
}

\begin{figure}[H]
\vspace{-0.4in}
\centering
\includegraphics[width=\linewidth]{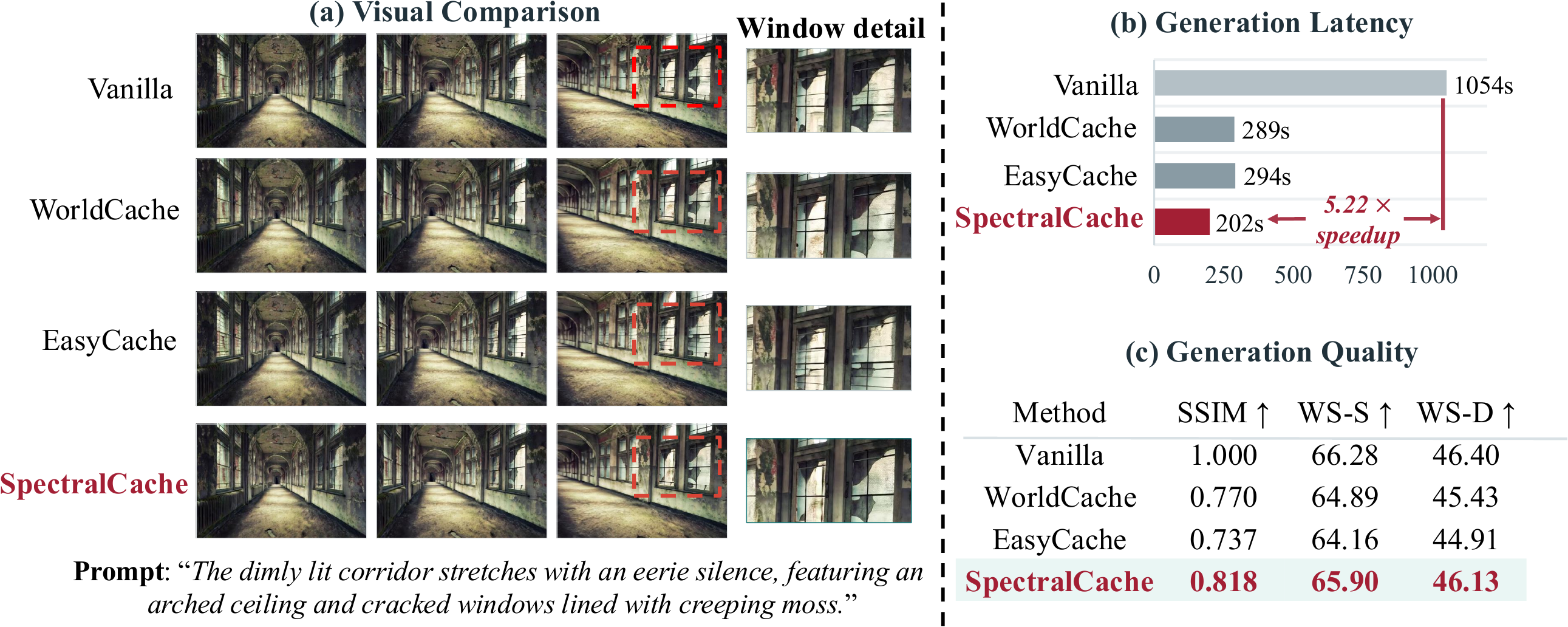}
%\vspace{-0.25in}
    \caption{
(a) Visual comparison, (b) latency comparison, and (c) quality comparison of SpectralCache with baselines on HunyuanWorld-Voyager-13B.
\textbf{SpectralCache achieves about $5.22\times$ acceleration while preserving high-fidelity video quality}.
% and fine-grained details}, such as the cracked windows highlighted by the red dashed boxes.
}
\label{subspace}
\vspace{-0.2in}
\end{figure}

\begin{abstract}

Diffusion-based world models enable high-quality interactive environment
generation but suffer from substantial inference overhead due to repeated
Transformer evaluations during denoising.
Existing caching methods mainly exploit temporal redundancy at the feature or
token level, leaving the underlying mathematical structure of diffusion
features largely unexplored.
In this work, we reveal that world-model features exhibit highly stable
singular subspaces across nearby denoising steps, while their singular values
follow predictable evolution patterns.
Building on this observation, we propose \textbf{SpectralCache}, a training-free
spectral caching framework that reuses stable singular subspaces and estimates
only low-dimensional singular values through linear extrapolation.
We further exploit the spectral consistency between neighboring
full-computation features to skip selected expensive backbone evaluations via
singular value scaling.
Extensive experiments on representative world models demonstrate
that SpectralCache consistently improves inference efficiency while preserving
generation quality.
On HunyuanWorld-Voyager-13B, SpectralCache achieves 5.22$\times$ acceleration
while maintaining a WorldScore of 65.90 for static scenes, substantially outperforming existing training-free caching methods in inference efficiency.

\end{abstract}

\section{Introduction}

% The ultimate goal of artificial intelligence is to achieve Artificial General Intelligence (AGI), enabling intelligent systems to perceive their surroundings, reason about diverse situations, and take appropriate actions in a human-like manner \cite{hafner2023mastering}. Despite the remarkable potential of LLMs toward AGI \cite{achiam2023gpt,bai2023qwen,chen2025empirical,llama2023llama,touvron2023llama}, their limited understanding and planning of dynamic physical environments remains a key challenge \cite{xiang2023language,dong2025probing}. Consequently, world models \cite{bruce2024genie,an2025embodied} have emerged as a promising direction to build more general-purpose intelligence \cite{bar2025navigation,liu2024sora}. Beyond common image or video generation, world models learn the underlying dynamics of environment and simulate its possible future evolution, providing predictive knowledge for long-horizon planning and decision-making \cite{agarwal2025cosmos,russell2025gaia}.
%  Among large-scale world models \cite{feng2026quantized,naveed2025comprehensive}, generation-driven diffusion-based world models have gained particular attention for 
%  superior generation of 
%  coherent and interactive virtual environments \cite{zhu2025aether,huang2025voyager}. 

 Diffusion-based world models enable controllable visual simulation by predicting future observations conditioned on scenes and actions \cite{bar2025navigation,zhu2025aether}. These capabilities support applications ranging from visual planning to explorable 3D scene generation \cite{bar2025navigation,huang2025voyager}. However, their iterative generation process requires repeated evaluations of large diffusion transformers (DiTs), incurring substantial computational cost and limiting responsiveness in interactive applications \cite{song2025hero,feng2026worldcache}. Accelerating world-model inference while preserving generation quality is therefore a key challenge for practical deployment.

% However, the substantial computational cost of diffusion transformers leads to slow inference, posing a major obstacle to the practical deployment of diffusion models in real-time applications \cite{ho2020denoising,lipman2022flow}. 

A wide range of diffusion acceleration methods have been proposed to improve  efficiency \cite{feng2025q,xi2025sparse,feng2025s,zhang2025sageattention}, including  refining ODE trajectories \cite{lu2022dpm,song2020denoising}, model pruning \cite{zhang2024effortless,guo2025mosaicdiff},  and quantization \cite{shang2023post,wu2025quantcache}. Among them, caching methods based on token structural characteristics have attracted increasing attention for diffusion acceleration \cite{chen2026forecast,selvaraju2024fora},  due to their training-free nature and ability to preserve superior generation quality by reusing or predicting intermediate representations.

% However, the substantial computational cost of diffusion transformers leads to slow inference, posing a major obstacle to the practical deployment of diffusion-based world models in real-time applications \cite{ho2020denoising,lipman2022flow}. Recently, a wide range of diffusion acceleration methods have been proposed to improve generation efficiency \cite{feng2025q,xi2025sparse,feng2025s,zhang2025sageattention}. These approaches broadly include: (1) \textbf{Optimization-based methods} such as refining ODE trajectories for fewer denoising steps can accelerate the diffusion process itself \cite{lu2022dpm,song2020denoising}, but they sometimes require additional training. (2) \textbf{Structure-based methods} can reduce computational costs through models modification techniques such as pruning \cite{zhang2024effortless,guo2025mosaicdiff}, quantization \cite{shang2023post,wu2025quantcache}, and caching methods based on token structural characteristics \cite{ma2024deepcache,zou2412accelerating}. With their training-free nature and ability to preserve better generation quality by reusing or predicting intermediate representations than models modification techniques, caching-based methods have attracted increasing attention for diffusion acceleration \cite{chen2026forecast,selvaraju2024fora}. 

However, most existing caching methods target conventional image and video diffusion models. They do not readily generalize to diffusion-based world models and direct adaptation often results in substantial quality degradation \cite{song2025hero}. Although recent studies have explored accelerating world model generation with caching methods \cite{song2025hero,feng2026worldcache}, they mainly focus on identifying world models' differences from conventional diffusion models and exploiting how  step-wise intermediate feature representations change for acceleration.
Different from these works, in this paper, we explore the evolvement of underlying mathematical structure and parameter-level characteristics throughout  generation, by using singular value decomposition (SVD).

We perform SVD on the intermediate DiT features throughout  world model generation  and analyze how their underlying subspaces and singular values evolve across generation steps. Our insightful analysis reveals that: (1) Singular subspaces remain highly stable with similar singular vectors across consecutive steps. (2) Singular values of  neighboring steps exhibit linear property, which enables accurate estimation of singular values through linear extrapolation or scaling.

% These observations suggest that, across consecutive denoising steps, future denoising features can be efficiently approximated by reusing preceding SVD subspaces and interpolating their singular values. Moreover, when accumulated caching errors necessitate a full computation, we can still reuse the previous full-computation subspace and approximate the current singular values through scaling, further reducing computational cost.

Building on these insights, we propose \textbf{SpectralCache}, a training-free caching framework for diffusion-based world models that efficiently accelerates video scene generation while preserving high generation quality. Specifically, across consecutive denoising steps, we approximate subsequent features by  interpolating singular values within the singular subspace of the preceding fully computed feature, thereby skipping redundant full computations. 
When a full computation step is required due to large enough  accumulation error \cite{feng2026worldcache}, instead of always using full computations, we alternatively adopt full computation and approximation of singular values through scaling,  yielding additional acceleration. Our comprehensive evaluation demonstrates that  SpectralCache achieves superior speedup with non-marginal improvements than SOTA world-model acceleration methods \cite{feng2026worldcache,song2025hero}. Meanwhile, it  leads to best generation quality compared with baselines due to highly accurate approximation. Our contributions are summarized as follows:
\begin{itemize}
    \item We propose \textbf{SpectralCache}, a training-free spectral feature caching framework for accelerating diffusion-based world models. SpectralCache reuses stable singular subspaces and estimates only low rank singular values through linear extrapolation for cache and scaling for skipping more full computation, substantially reducing expensive overhead while preserving generation quality.

    \item We reveal several interesting spectral structures in world-model features: they exhibit strong low-rank structure, while their dominant singular subspaces remain highly stable across both consecutive denoising steps and nearby cache-triggered full computation steps. Moreover, their singular values exhibit predictable linear extrapolation and approximately constant scaling patterns in these two settings, respectively, providing a new spectral perspective for understanding and exploiting temporal redundancy in diffusion-based world models.

    \item Across two representative world models, HunyuanWorld-Voyager-13B and Aether-5B, SpectralCache consistently outperforms existing training-free caching methods in both generation quality and inference efficiency. On HunyuanWorld-Voyager-13B, SpectralCache achieves a 5.22$\times$ speedup while maintaining WorldScores of 65.90 and 46.13 for static and dynamic scenes, respectively, compared with 3.65$\times$ speedup and WorldScores of 64.89 and 45.43 achieved by WorldCache.
\end{itemize}

%These observations suggest that consecutive denoising steps can be efficiently approximated via singular value extrapolation within subspaces of preceding steps, while steps where caching is no longer applicable can be accelerated through subspace reuse and singular value scaling. Building on these insights, we propose XXX, a training-free caching framework for diffusion-based world models that efficiently accelerates video scene generation while preserving high generation quality. Specifically, across consecutive denoising steps, we approximate subsequent features by interpolating their singular values within the subspace of the preceding fully computed feature, thereby skipping redundant full computations. When a full computation is required according to existing cache error criteria \cite{feng2026worldcache}, we further reuse the SVD subspace from the previous true full computation and scale its singular values to approximate the current feature, yielding additional acceleration. Since our method interpolates only a few singular value ranks during cached steps and further skips steps that would otherwise require full computation, it achieves greater speedup than existing world-model acceleration methods \cite{feng2026worldcache,song2025hero} while maintaining comparable generation quality. Our contributions are summarized as follows:

% \textcolor{red}{@please list the contributions.}

\section{Related Works}

%\textcolor{red}{related work can be shrinked}

\textbf{Generation-driven Diffusion World Models.} World models seek to learn predictive representations of the environment that support future-state anticipation, planning, and decision-making \cite{hafner2019dream,hafner2023mastering}. %Existing approaches can be broadly grouped into two paradigms. 
Policy-centric world models \cite{hafner2019dream,hafner2023mastering,robine2023transformer,wu2023daydreamer} acquire environmental dynamics through agent-environment interaction, where action-conditioned observations and feedback are progressively incorporated into an internal model for subsequent planning and control. 
%Their key challenge is to learn expressive yet compact dynamics representations that can reliably support downstream decisions \cite{hou2026world}.
In contrast, generative world models \cite{agarwal2025cosmos,bar2025navigation,russell2025gaia,zhu2025aether} learn environmental dynamics from large-scale visual or video data and simulate plausible future scene evolution with diffusion method. %Such models can synthesize temporally coherent and interactive environments, with Aether \citet{zhu2025aether} representing a recent example that extends a video foundation model to multimodal world simulation. 

\textbf{Model-level Diffusion Acceleration.}
Despite their strong generative capabilities \cite{yang2023diffusion,croitoru2023diffusion}, diffusion models still incur substantial inference costs, motivating various acceleration strategies. Optimization-based methods primarily modify the sampling process without altering the underlying denoising network, using deterministic sampling or advanced high-order solvers to reduce the number or cost of denoising steps \cite{guo2024gradient,lu2022dpm,song2020denoising}. Recent studies further optimize ODE trajectories through distillation \cite{luo2024one,yan2024perflow,yin2024one}, enabling generation with substantially fewer sampling steps, but often requiring additional optimization or training \cite{salimans2022progressive}. In contrast, model-structural acceleration methods directly reduces the computational complexity of the denoising network itself, such as compressing model parameters through quantization \cite{shang2023post,wu2025quantcache} or removing redundant structures through pruning \cite{zhang2024effortless,guo2025mosaicdiff}. 
%However, reducing model parameters or modifying model structures inevitably introduces approximation errors, which may degrade generation quality under aggressive compression~\cite{he2023ptqd,chen2025q}. This limitation has motivated recent studies to shift their focus toward accelerating generation at the feature-token level, which is also the direction explored in our work.

\textbf{Diffusion Acceleration from Representation Cache.}
Caching has recently emerged as an effective training-free approach for  accelerating diffusion inference while largely preserving generation quality \cite{kahatapitiya2025adaptive,ma2024deepcache}. Adaptive token caching  selectively reuses tokens based on their importance or variation while recomputing the remaining tokens 
\cite{zou2024accelerating,zheng2025compute}. Forecasting-based methods instead predict the evolution of intermediate representations to skip subsequent full computations across denoising steps \cite{liu2025reusing,chen2026forecast}.
However, most existing caching methods are designed for conventional image or video generation and often suffer from quality degradation when directly applied to world models with complex hierarchical feature dynamics \cite{song2025hero}. 

\section{Deep Insights with Singular Value Decomposition}
\label{motivation}

\begin{figure}[b]
    \centering
    \vspace{-0.1in}
    \includegraphics[width=0.9\linewidth]{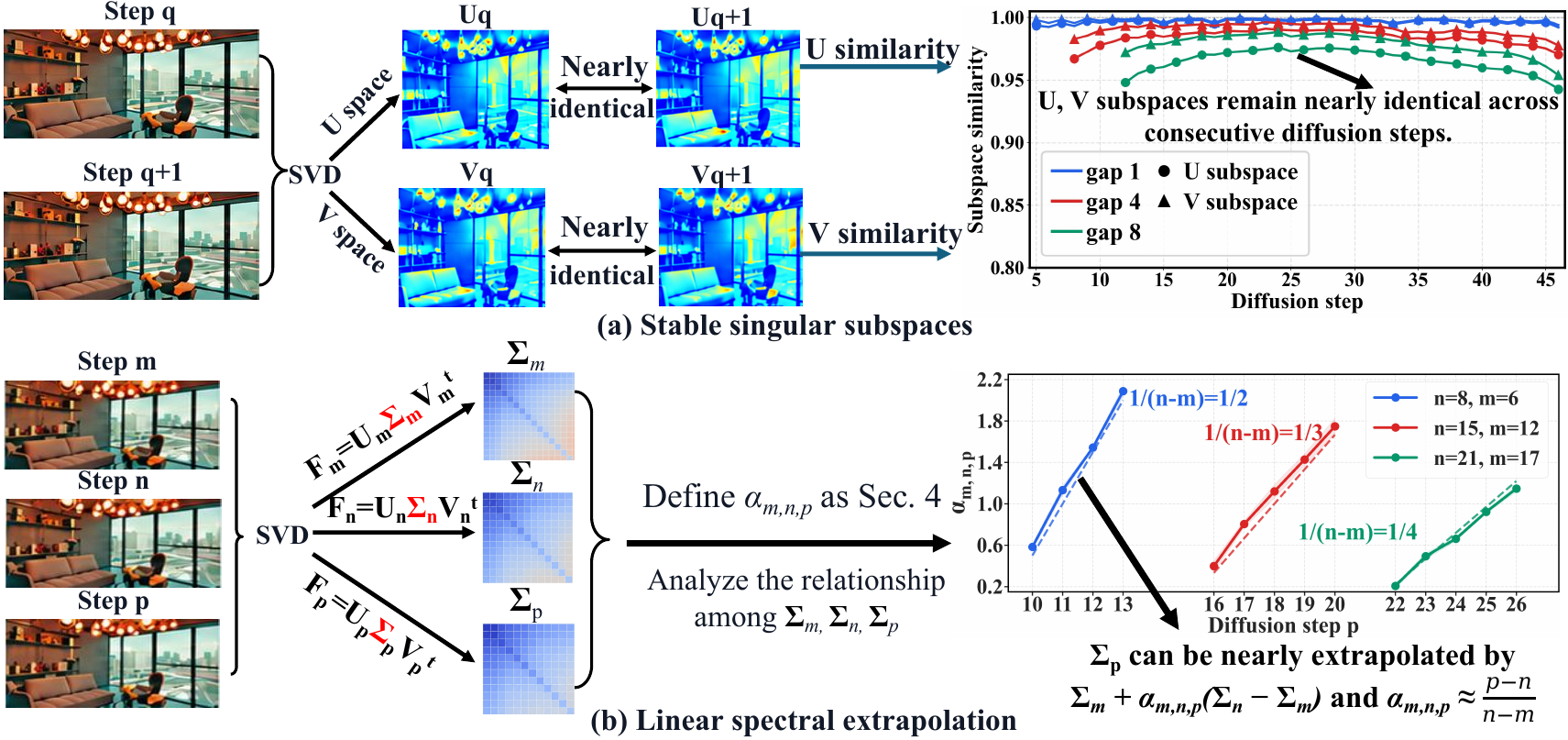}
    \vspace{-0.1in}
    \caption{
    Spectral dynamics across denoising steps in HunyuanWorld-Voyager.
(a) The dominant singular subspaces remain highly stable across consecutive denoising steps.
(b) Singular values exhibit a consistent linear extrapolation pattern across denoising steps.
% \textcolor{red}{(title is too long. many information are already covered in contexts.)}
    % \textcolor{red}{(figure 2 and 3 take almost one page. maybe move the whole figure to appendix, and only keep the academic plots. maybe do not need the explanation parts, it is easy to understand singular vector replacement due to cos similarity and already discussed in context. no need to explain again with figures. the explanation parts take a lot of space, but provide little new information.)}
    }
% \caption{
% Spectral dynamics across denoising steps in HunyuanWorld-Voyager.
% (a) The dominant singular subspaces remain highly stable across nearby steps.
% (b) Singular values exhibit a consistent linear extrapolation pattern across denoising steps.
% }
    \label{fig:uvsapce}
    \vspace{-0.25in}
\end{figure}

% \begin{figure}
% %\vspace{-0.1in}
% \centering
% \includegraphics[width=\linewidth]{Figure/subspace.pdf}
% %\vspace{-0.25in}
%     \caption{
% Singular subspace stability across denoising steps in HunyuanWorld-Voyager.
% The curves show cosine similarity between $U$ and $V$ subspaces at steps $t$ and $t-k$, where \emph{gap} $k$ is their denoising step distance.
% Both subspaces remain highly aligned for gaps up to eight steps. \textcolor{red}{(larger fonts in figures)}
% }
% \label{subspace}
% %\vspace{-0.1in}
% \end{figure}

% \begin{wrapfigure}{r}{0.55\textwidth}
%     \centering
%     \vspace{-8pt}
%     \includegraphics[width=\linewidth]{Figure/extra.pdf}
%     \caption{
%     The proportionality coefficient $\alpha_{m,n,p}$ closely follows
%     $\frac{p-n}{n-m}$ for each subsequent step $p>n$, with a constant slope
%     $\frac{1}{n-m}$. Solid and dashed lines denote the measured values and
%     ideal linear trends, respectively.
%     }
%     \label{fig:layeralign}
%     \vspace{-8pt}
% \end{wrapfigure}

\textbf{Low-rank spectral state of visual features.} 
% \textcolor{red}{(maybe not mention 'trigger' in this section. We formally introduce the full algorithm with 'trigger' in section 4. So in section 3, readers are not supposed to understand 'trigger'. So I just say nearby steps rather than 'triggered steps'.)}
We adopt SVD to analyze visual features. 
The $t^{th}$ denoising step visual feature $\mathbf{F}_t\in \mathbb{R}^{f \times h \times w \times D}$ from final-layer of DiT backbone (with  $f$ frames, height  $h$, and  width  $w$) can be  transformed to  $\mathbf{X}_t = \operatorname{mat}(\mathbf{F}_t) \in \mathbb{R}^{M \times D}$, where $M=f \times h \times w$ denotes the number of visual tokens and $D$ is the channel dimension. We can perform SVD on $\mathbf{X}_t$ as 
\begin{equation}
    \mathbf{X}_t
    = \mathbf{U}_t \mathbf{\Sigma}_t \mathbf{V}_t^{\top}
    + \mathbf{R}_t,
    \label{eq:truncated_svd_state}
\end{equation}
where $\mathbf{U}_t \in \mathbb{R}^{M \times r}$ and
$\mathbf{V}_t \in \mathbb{R}^{D \times r}$ are the leading singular
subspaces, $\mathbf{\Sigma}_t \in \mathbb{R}^{r \times r}$ is diagonal, and
$\mathbf{R}_t$ is the truncated residual. We call
$\mathcal{Q}_t
    = \{\mathbf{U}_t, \mathbf{\Sigma}_t,
       \mathbf{V}_t, \mathbf{R}_t\}$
the spectral state at $t$-th step. 

Our key observations with SVD provides deep insights for features during diffusion. Specifically, we perform SVD as $\mathbf{X}_t = \mathbf{U}_t \mathbf{\Sigma}_t \mathbf{V}_t^\top$ for the visual feature $\mathbf{F}_t$ generated at the $t^{th}$ denoising step.
%where $\mathbf{U}_t$ and $\mathbf{V}_t$ denote the left and right singular subspaces, respectively, and $\mathbf{\Sigma}_t$ contains the corresponding singular values. 
By carefully analyzing the singular vectors and values, we can make the following key observations.

\textbf{Denoising features are low-rank.}  We first observe that
visual features exhibit a strong low-rank structure, with the top 128 singular values capturing  96\% of the total feature energy on average.

\textbf{Singular subspaces remain highly stable with similar singular vectors.}
To characterize how these singular subspaces evolve throughout denoising, we measure their alignment between different steps with principal-angle similarity. Given
$\mathbf{X}_t=\mathbf{U}_t \mathbf{\Sigma}_t \mathbf{V}_t^\top$ and
$\mathbf{X}_{t+k}=\mathbf{U}_{t+k}\mathbf{\Sigma}_{t+k}\mathbf{V}_{t+k}^\top$, we compute
\begin{equation}
\begin{aligned}
\cos(\theta_i^{(\mathbf{U},t,k)})
&= \left[\mathbf{U}_t^\top \mathbf{U}_{t+k}\right]_{i,i},
\qquad
\cos(\theta_i^{(\mathbf{V},t,k)})
= \left[\mathbf{V}_t^\top \mathbf{V}_{t+k}\right]_{i,i}, \\
\overline{\cos(\theta^{(\mathbf{U},t,k)})}
&= \frac{1}{a}\sum_{i=1}^{a}\cos(\theta_i^{(\mathbf{U},t,k)}),
\qquad
\overline{\cos(\theta^{(\mathbf{V},t,k)})}
= \frac{1}{a}\sum_{i=1}^{a}\cos(\theta_i^{(\mathbf{V},t,k)}).
\end{aligned}
\label{eq:subspace_similarity}
\end{equation}
$[\cdot]_{i,i}$ denotes the $i^{th}$ diagonal element of the matrix. % \textcolor{red}{(check, equation (9) also uses $\sigma_i$, maybe use another expression}.
%$\sigma_i(\cdot)$ denotes the $i$-th singular value \textcolor{red}{(check, equation (9) also uses $\sigma_i$, maybe use another expression}, 
$\cos(\theta_i^{(\mathbf{U},t,k)})$ is the cosine similarity of the $i^{th}$ singular vectors between different steps, and $a$ denotes the number of evaluated singular directions. 
A higher average cosine similarity indicates stronger subspace consistency across denoising steps.

We plot the cosine similarity to preceding steps at each diffusion step under various gap $k$ configurations. As shown in Figure~\ref{fig:uvsapce}(a), the leading $\mathbf{U}$ and $\mathbf{V}$ subspaces remain highly aligned  at temporal gap of up to eight denoising steps, with a cosine similarity mostly above 0.95.

\begin{figure}[t]
    \centering
    \vspace{-0.1in}
    \includegraphics[width=0.9\linewidth]{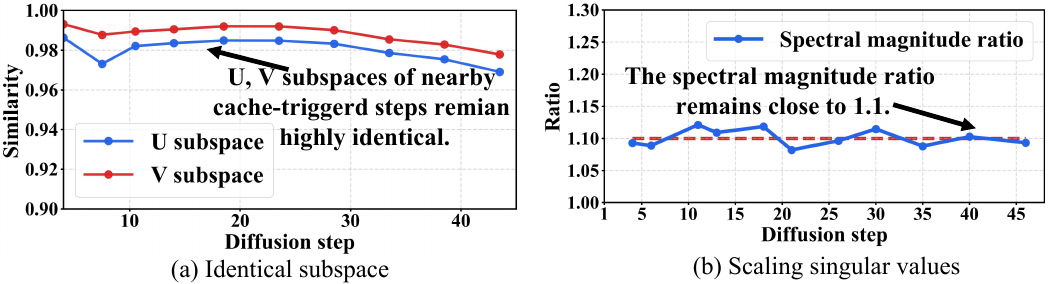}
    \vspace{-0.1in}
    \caption{
    Spectral consistency across nearby  full-computation steps.
(a) The dominant singular subspaces remain highly consistent.
(b) The spectral magnitude ratio remains approximately constant.
% \textcolor{red}{(figure 2 and 3 take almost one page. maybe move the whole figure to appendix, and only keep the academic plots. maybe do not need the explanation parts, it is easy to understand replacement due to cos similarity and already discussed in context. no need to explain again with figures.)} 
% \textcolor{blue}{(a) can be removed. We can put the lower part of  figure to the curve figure's right. We should use math format for each notation instead of text. Background color (blue) of step feature, U, V can be removed.}
    }
    \label{fig:space_singular}
    \vspace{-0.25in}
\end{figure}

\textbf{Singular values  exhibit a linear extrapolation pattern.}  We observe that the singular values  exhibit an approximately linear extrapolation pattern with respect to those from preceding diffusion steps.
Specifically, we denote the singular values of the fully computed features at steps $m$, $n$, and $p$ as %\textcolor{red}{(check, they should be vectors, but the expression is matrix as in equation 3)} 
$\operatorname{diag}(\mathbf{\Sigma}_m)$, $\operatorname{diag}(\mathbf{\Sigma}_n)$, and $\operatorname{diag}(\mathbf{\Sigma}_p)$ ($m<n<p$), respectively. We compute
$\alpha_{m,n,p}
=
\frac{
\left\langle
\operatorname{diag}(\mathbf{\Sigma}_p)-\operatorname{diag}(\mathbf{\Sigma}_n),\,
\operatorname{diag}(\mathbf{\Sigma}_n)-\operatorname{diag}(\mathbf{\Sigma}_m)
\right\rangle
}{
\left\|\operatorname{diag}(\mathbf{\Sigma}_n)-\operatorname{diag}(\mathbf{\Sigma}_m)\right\|_2^2
}.$
We randomly sample a few combinations of $(m,n)$, and plot the value of  $\alpha_{m,n,p}$ with respect to $p$. As shown in Fig.~\ref{fig:uvsapce}(b),  under fixed $m$ and $n$, $\alpha_{m,n,p}$ increases almost linearly as $p$ increases, with a slope close to $\frac{1}{n-m}$, i.e.,
$\alpha_{m,n,p}
\approx
\frac{p-n}{n-m}.$
It suggests that  singular values of subsequent features can be extrapolated from those of  two preceding  steps:
\begin{equation}
\operatorname{diag}(\mathbf{\hat\Sigma}_p)
\approx
\operatorname{diag}(\mathbf{\Sigma}_n)
+
\frac{p-n}{n-m}
\left(\operatorname{diag}(\mathbf{\Sigma}_n)-\operatorname{diag}(\mathbf{\Sigma}_m)\right),
\qquad p>n>m.
\end{equation}
This approximation empirically introduces only a $0.71\%$ error relative to the corresponding fully computed features throughout  denoising, motivating us to approximate subsequent features by  extrapolating  their preceding singular values, thereby reducing computational overhead by caching.

\textbf{Singular values can be approximated with a constant scaling factor.} 
% We further investigate the relationship between the feature singular values at two nearby computation steps.
% We further investigate the relationship between feature singular values at two nearby full computation steps separated by several cache intervals.
We further investigate the relationship between feature singular values across nearby full computation steps separated by several steps.
%As shown in Figure~\ref{fig:space_singular}(a), the $U$ and $V$ subspaces remain highly consistent between the current triggered full computation step $t$ and the most recent preceding full computation step $q$.
%Given such stable singular subspaces, we further examine the variation in their corresponding singular values.
Given the nearly identical $\textbf{U},\textbf{V}$ subspaces between the two steps (Figure \ref{fig:space_singular}(a)) 
% we further examine the ratio of their singular value norms
% $\frac{|\operatorname{diag}(\mathbf{\Sigma}_t)|_2}{|\operatorname{diag}(\mathbf{\Sigma}_q)|_2}$.
, we examine the ratio of singular value norms $\frac{\|\operatorname{diag}(\mathbf{\Sigma}_t) \|_2}{\| \operatorname{diag}(\mathbf{\Sigma}_q)\|_2}  $  %\textcolor{red}{(check, maybe $\| \operatorname{diag}(\Sigma_t) \|_2  / \| \operatorname{diag}(\Sigma_q)\|_2$? vector is hard to divide)} 
between two nearby  steps $t$ and $q$, as shown in Figure~\ref{fig:space_singular}(b). The ratio $\frac{\|\operatorname{diag}(\mathbf{\Sigma}_t) \|_2}{\| \operatorname{diag}(\mathbf{\Sigma}_q)\|_2}  $ 
remains nearly constant, suggesting that the singular values of neighboring features differ mainly by a constant scaling factor.  
Motivated by this, we can approximate the current feature by simply rescaling previous step's singular values, thereby avoiding  expensive  computations.

% \begin{figure}[t]
% \centering

% \begin{minipage}[t]{0.48\linewidth}
%     \centering
%     \includegraphics[width=\linewidth]{Figure/uvspace.pdf}
%     \caption{
%     The $U$ and $V$ subspaces between two nearby triggered full computation steps
% $t$ and $q$ exhibit high consistency.
% $\bullet$ denotes triggered full-computation steps.
%     }
%     \label{fig:uvspace}
% \end{minipage}
% \hfill
% \begin{minipage}[t]{0.48\linewidth}
%     \centering
%     \includegraphics[width=\linewidth]{Figure/ratio.pdf}
%     \caption{
%     The singular value magnitude ratio between two nearby triggered
%     full computation steps,
%     $\frac{\|\operatorname{diag}(\mathbf{\Sigma}_t) \|_2}{\| \operatorname{diag}(\mathbf{\Sigma}_q)\|_2}  $,
%     remains approximately constant.
%     }
%     \label{fig:ratio}
% \end{minipage}
% \vspace{-0.2in}
% \end{figure}

%\input{sections/motivation_old}

%\input{sections/method_old}

\section{Method}
\label{sec:method}

Based on our key observations and deep insights for the singular vectors and values of denoising features across different steps, we propose a spectral feature caching framework  to accelerate diffusion-based world models. The core idea is to approximate diffusion features  by reusing  cached results of preceding full computation steps, 
%the features at certain diffusion steps by reusing  cached results of preceding full computation steps, 
thus skipping computations and achieving acceleration. 
%Specifically, the approximated features are obtained through SVD, with singular vectors from preceding fully computation step due to high similarity, and singular values from linear extrapolation or scaling of preceding steps due to linear property. 
Specifically, we define two kinds of diffusion steps: full computation step and approximation step. In full computation step, we go through the normal full diffusion computations and cache the singular vectors/values. In the approximation step, we do not perform the   diffusion computations. Instead, the approximated features reuse the highly consistent singular subspaces from preceding full computation steps, while the singular values are estimated through linear extrapolation   or scaling, following their respective spectral evolution patterns based on  our observations and insights.
%while the singular values are estimated through linear extrapolation at cache steps or scaling at cache-triggered full computation steps, following their respective spectral evolution patterns. %The details are explained below. 

%\textcolor{red}{(determine the name for some components, such as full computation step v.s. computation step, or singular value or singular-value, then use the same consistent name throughout the paper. try not use different names for the same thing.)}

\subsection{Full Computation Step}

At a full computation step $t$, we perform a complete DiT backbone evaluation
to obtain the final layer visual feature $\mathbf{F}_t$ and reshape it into
$\mathbf{X}_t=\operatorname{mat}(\mathbf{F}_t)\in\mathbb{R}^{M\times D}$.  Then we compute its rank-$r$ spectral decomposition for potential subsequent cache steps:
\begin{equation}
\mathbf{X}_t
=
\mathbf{U}_t^r
\mathbf{\Sigma}_t^r
(\mathbf{V}_t^r)^\top
+
\mathbf{R}_t,
\end{equation}
where $\mathbf{U}_t^r$, $\mathbf{\Sigma}_t^r$, and $\mathbf{V}_t^r$ denote the top-$r$ spectral subspaces and singular values, and $\mathbf{R}_t$ is the truncated residual. Based on our deep insights, we choose $r=128$ since it captures 96\% feature energy.
We store the resulting spectral state
$\mathcal{Q}_t=\{\mathbf{U}_t^r,\mathbf{\Sigma}_t^r,
\mathbf{V}_t^r,\mathbf{R}_t\}$ %as a reference 
for subsequent steps.
The spectral states of only two most recent full computation steps are cached for reuse.% rather than all computation steps.
%We retain the spectral states from the two most recent computation steps for subsequent cache steps. 
%Once a new  computation step is triggered, the older state  of the cached two steps is discarded and replaced by the newly computed state.

\begin{figure}[t]
    \centering
    \vspace{-0.1in}
    \includegraphics[width=0.99\linewidth]{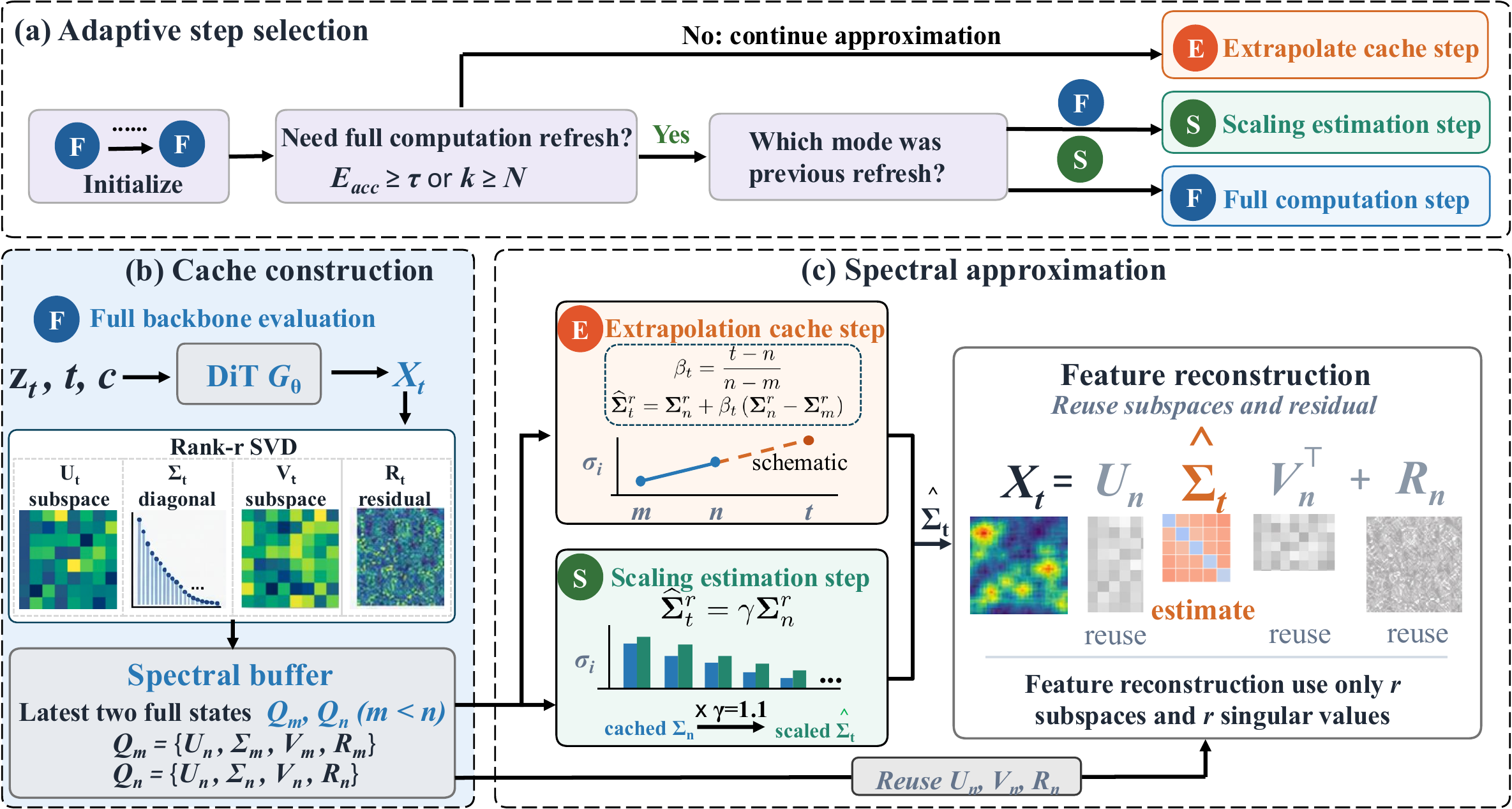}
        % {Figure/all.pdf}
    \vspace{-0.1in}
    \caption{
    Overview of SpectralCache.
    % \textcolor{red}{This figure should be mentioned somewhere in the paper}
    }
    \label{fig:pipeline}
    \vspace{-0.2in}
\end{figure}

\subsection{Approximation Steps}
\label{sec:approximation}

At an approximation step, we bypass the full DiT backbone computation and reconstruct the  visual feature by reusing the saved singular subspaces and residual while estimating only the dominant singular values.  Specifically, the current visual feature at step $t$ is approximated as
\begin{equation}
    \widehat{\mathbf{X}}_{t}
    =
    \mathbf{U}_{n}^{r}
    \widehat{\mathbf{\Sigma}}_{t}^{r}
    (\mathbf{V}_{n}^{r})^{\top}
    +
    \mathbf{R}_{n}.
    \label{eq:feature_extrapolation}
\end{equation}
where $\mathbf{U}_{n}^{r}$ and $\mathbf{V}_{n}^{r}$ are spectral subspaces of 
the most recent full computation step.
Since the leading singular subspaces remain highly stable across nearby denoising steps as demonstrated in Section~\ref{motivation}, we directly reuse $\mathbf{U}_{n}^{r}$ and $\mathbf{V}_{n}^{r}$.
The residual $\mathbf{R}_{n}$ is also directly reused since it accounts for only
about $3\%$ of the total feature energy on average throughout generation.

$\widehat{\mathbf{\Sigma}}_{t}^{r}$ is estimated from singular values of previous full computation steps. We employ two estimation  strategies according to the current caching state: linear extrapolation for consecutive cache steps and singular value scaling when a full computation step is triggered but can be safely skipped.
Thus, each approximation step requires updating only $r$ singular values while reusing
other spectral components from  the most recent full computation step,
avoiding expensive DiT backbone computations.

\subsubsection{Singular Value Cache through Linear Extrapolation}
\label{sec:sv_extrapolation}

%For consecutive cache steps, we approximate the current visual feature by reusing the singular subspaces from the most recent two full computation step and estimating only its singular values. 
We can estimate singular values through linear extrapolation for consecutive denoising steps. Specifically,  let $m<n<t$ denote denoising steps, where $m$ and $n$ are the two most recent full computation steps and $t$ is the current  step to be approximated. Based on  approximately linear evolution of singular values observed in Section~\ref{motivation}, 
we can extrapolate  singular values of  current step $t$ from step $m$ and $n$: %from two most recent full computation steps: 
\begin{equation}
    \widehat{\mathbf{\Sigma}}_{t}^{r}
    =
    \mathbf{\Sigma}_{n}^{r}
    +
    \beta_t \left( \mathbf{\Sigma}_{n}^{r}
    -
    \mathbf{\Sigma}_{m}^{r} \right),
    \qquad
    \beta_t
    =
    \frac{t-n}{n-m},
    \label{eq:sv_extrapolation}
\end{equation}
% Equivalently, each singular value is estimated as
% \begin{equation}
%     \hat{\sigma}_{t}^{i}
%     =
%     \sigma_{n}^{i}
%     +
%     \frac{t-n}{n-m}
%     \left(
%     \sigma_{n}^{i}-\sigma_{m}^{i}
%     \right),
%     \qquad i=1,\ldots,r.
%     \label{eq:individual_sv_extrapolation}
% \end{equation}
%As shown in Equation~\eqref{eq:sv_extrapolation}, 
As shown above, $\widehat{\mathbf{\Sigma}}_{t}^{r}$ is mainly determined by the step distance  of $m,n,t$, and the difference of cached visual  features $\mathbf{\Sigma}_{n}^{r}$ and $\mathbf{\Sigma}_{m}^{r}$.  Our analysis shows that  this approximation empirically introduces only a 0.71\% error relative to the corresponding ground truth fully computed features throughout denoising, showing that this approximation is highly accurate with minor errors.

% We then extrapolate the singular values at the current step according to its
% relative denoising-step distance:
% \begin{equation}
%     \beta_t
%     =
%     \frac{t-n}{n-m},
%     \qquad
%     \widehat{\mathbf{\Sigma}}_{t}^{r}
%     =
%     \mathbf{\Sigma}_{n}^{r}
%     +
%     \beta_t\Delta\mathbf{\Sigma}_{m,n}^{r},
%     \label{eq:sv_extrapolation}
% \end{equation}
% where $\beta_t$ controls the extrapolation magnitude according to the distance
% from the most recent full-computation step, and
% $\widehat{\mathbf{\Sigma}}_{t}^{r}$ denotes the estimated top-$r$ singular
% values at step $t$. 

\subsubsection{Singular Value Estimation through Scaling}
\label{sec:sv_scaling}
 
We propose another way to estimate singular values. As observed in Section~\ref{motivation}, 
%the dominant singular subspaces and singular-value directions remain highly consistent, while 
the singular value magnitudes differ primarily by a constant scaling factor between nearby cache-triggered full computation steps. Thus, 
%we estimate singular values through scaling. Specifically, 
%let $t$ denote the current approximation step,
%requiring singular value  estimation,  
% and $n<t$ denote its most recent full computation step. 
%Based on the cached spectral state
% $\mathcal{Q}_n=
% \{\mathbf{U}_n^r,\mathbf{\Sigma}_n^r,
% \mathbf{V}_n^r,\mathbf{R}_n\}$,
we can estimate singular values at full computaion step $t$ as
\begin{equation}
    \widehat{\mathbf{\Sigma}}_t^r
    =
    \gamma\mathbf{\Sigma}_n^r,
    \label{eq:sv_scaling}
\end{equation}
where $\gamma$ is a fixed scaling factor controlling the magnitude adjustment, and $n<t$ denotes its most recent full computation step.
We use $\gamma=1.1$ (measured in Section~\ref{motivation}) for all main experiments.  %\textcolor{red}{(observation only shows that the magnitude can be scaled, does not show that the dimension is the same. but here, we directly use the same direction. I am not sure  Maybe we can say that the first estimation method is more accurate, and this second estimation is not that accurate due to direction. Thus we mainly use the first estimation in algorithm, and use the second estimation less frequently.)}

\subsection{Approximation Error Accumulation to Trigger Full Computation}
\label{sec:error_accumulation}

For each diffusion step, we need to determine whether it is a full computation step or an approximation step. To complete this, we adopt an approximation error accumulation strategy to trigger full computation steps. 
In detail, although our efficient approximation steps introduces minor errors in each step, consecutive approximations across multiple denoising  steps  may gradually accumulate the estimation errors. When this accumulation error is large enough, we stop approximation and a full computation step is triggered to ensure high quality generation. 

%\textcolor{red}{(the following part mainly follows existing work, it is not our contribution, maybe move to appendix or keep one   core equation, no need to show two equations.)}
% Specifically, following \cite{feng2026worldcache}, we monitor the accumulated approximation error of chaotic tokens, whose features exhibit less predictable temporal token variations.  For each chaotic token $q\in\mathcal{T}_{\mathrm{chaotic}}$, we compute its curvature-weighted approximation error as
% \begin{equation}
% e_q(t)
% =
% \kappa_q
% \left\|
% \frac{\mathbf{x}_{t,q}}{\|\mathbf{x}_{t,q}\|_2}
% -
% \frac{\mathbf{x}_{t+1,q}}{\|\mathbf{x}_{t+1,q}\|_2}
% \right\|_2, \ \ \ \ 
% \kappa_q
% =
% \left\|
% \mathbf{x}_{t+1,q}
% -
% 2\mathbf{x}_{t,q}
% +
% \mathbf{x}_{t-1,q}
% \right\|_2.
% \label{eq:token_error}
% \end{equation}

Following~\cite{feng2026worldcache}, we monitor approximation error over chaotic tokens to determine when cached feature approximation should be terminated. Specifically, we accumulate the average curvature-weighted feature variation across consecutive extrapolation-based cache steps: 
\begin{equation} 
E_{\mathrm{acc}} \leftarrow E_{\mathrm{acc}} + \frac{1}{|\mathcal{T}_{\mathrm{chaotic}}|} \sum_{q\in\mathcal{T}_{\mathrm{chaotic}}} e_q(t), k \leftarrow k+1.\label{eq:accumulated_error} 
\end{equation} 
where $\mathcal{T}_{\mathrm{chaotic}}$ denotes the set of tokens exhibiting less predictable temporal feature variations, $k$ denotes the total number of cache steps performed, and $e_q(t)$ denotes the curvature-weighted approximation error of the $q$-th chaotic token following~\cite{feng2026worldcache}.
In general, if the accumulated error is smaller than a threshold $E_{\mathrm{acc}}<\tau$ and the maximum cache interval $N$ is not reached, the diffusion goes through the approximation step for efficiency. Once $E_{\mathrm{acc}}\geq\tau$ or the cache interval $k$ reaches $N$ \cite{feng2026worldcache}, a full computation step is triggered to prevent excessive approximation error.

\subsection{Overall Pipeline}

We show our overall pipeline in Figure \ref{fig:pipeline} with detailed algorithm in Appendix \ref{app:algorithm}. After two initial   full computation steps, it goes through consecutive approximation steps with linear extrapolation for singular values, meanwhile recording the accumulated error $E_{\mathrm{acc}}$ \cite{feng2026worldcache}.  
If  $E_{\mathrm{acc}}>\tau$ or the maximum cache interval is reached, ideally a full computation step is triggered for a feature refresh to reduce error. In practice,  instead of always executing the expensive  full computation step when triggered, we alternatively adopt full computation or approximation with scaling of singular values, yielding additional acceleration. 
Specifically,  if the previous triggered step is full computation, we use approximation step with scaling to estimate singular values for the current triggered step. Otherwise if the previous triggered step is approximation step, we adopt a full computation step.
Then after the triggered step, it continues to approximation steps with linear extrapolation until triggered again by accumulation error.

\section{Experiments}

\subsection{Experimental Settings}

\textbf{Models.} We evaluate our method on two SOTA mainstream diffusion based world models: HunyuanVoyager-13B \cite{huang2025voyager} and Aether-5B \cite{zhu2025aether}. Both models take images, text, and camera trajectories as inputs and jointly generate RGB videos and corresponding depth maps. These models take an initial image, an instruction, and camera poses as conditioning inputs to generate 3D videos comprising both RGB frames and scene depth.

\textbf{Datasets and Evaluation Metric.} For the video generation datasets and evaluation metrics, we follow the experimental settings of prior works \cite{song2025hero,feng2026worldcache}. For generation quality evaluation, we similarly adopt WorldScore \cite{duan2025worldscore}, a comprehensive benchmark that assesses world generation in terms of both controllability and visual quality. Additionally, we evaluate the perceptual consistency of the generated videos by measuring their discrepancy from the no-cache baseline using PSNR, SSIM \cite{wang2004image}, and LPIPS \cite{zhang2018unreasonable}. Following the evaluation protocol of existing work \cite{song2025hero}, we also evaluate the 3D reconstruction metrics of generation videos with Sintel dataset \cite{butler2012naturalistic}, which contains AbsRel, $\delta_1,\delta_2$, ATE, and
RPE-t/RPE-r. We also report speedup and peak memory.

% \textbf{Baselines.} \textcolor{red}{(due to space limit, there are many baselines and maybe only mention their names.)}
% We compare our method against several effective training-free diffusion model caching methods. DuCa \cite{zou2412accelerating} alternates aggressive and conservative feature caching to reduce redundant computation and caching errors. ToCa \cite{zou2024accelerating} adaptively caches less-sensitive tokens while recomputing critical tokens to reduce redundant computation. TaylorSeer \cite{liu2025reusing} predicts future features using Taylor-series expansion and higher-order finite differences instead of directly reusing cached features. HiCache \cite{feng2026hicache} replaces Taylor-based feature forecasting with scaled Hermite polynomials for more accurate and stable feature prediction. TeaCache \cite{liu2025timestep} uses timestep-modulated input differences to adaptively determine when to cache and reuse model outputs. EasyCache \cite{zhou2025less} dynamically reuses transformation vectors based on runtime feature stability, without offline profiling or training. HERO \cite{song2025hero} uses patch-wise token refresh in shallow layers and linear feature extrapolation in deeper layers to accelerate world models. WorldCache \cite{feng2026worldcache} uses curvature-guided heterogeneous token prediction and chaotic-prioritized adaptive skipping to efficiently cache diffusion world models.

\textbf{Baselines.}
We compare our method against several effective training-free diffusion model caching methods, including DuCa~\cite{zou2412accelerating}, ToCa~\cite{zou2024accelerating}, TaylorSeer~\cite{liu2025reusing}, HiCache~\cite{feng2026hicache}, TeaCache~\cite{liu2025timestep}, EasyCache~\cite{zhou2025less}, HERO~\cite{song2025hero}, and WorldCache~\cite{feng2026worldcache}.

\textbf{Implementation Details.}
All experiments are conducted on two NVIDIA Pro 6000 96GB GPUs. We use a rank of $128$ for all experiments and set $\gamma = 1.1$.
%the singular value scaling factor to $1.1$ for the main results. 
%For the residual component, we reuse the residual from the preceding step for both singular value extrapolation and scaling. 
For other settings, we follow prior works~\cite{feng2026worldcache,zhu2025aether,song2025hero} to make a fair comparison (see Appendix \ref{experiments}). 
%For the settings, including the total number of denoising steps, number of warm-up steps, accumulated error threshold, proportion of chaotic tokens, and benchmark evaluation protocols, we follow prior works~\cite{feng2026worldcache,zhu2025aether,song2025hero}

\begin{table}[ht]

\centering
% \small
% \setlength{\tabcolsep}{5pt}
% \renewcommand{\arraystretch}{1.15}
\vspace{-0.1in}
\begin{adjustbox}{width=\columnwidth}
\begin{tabular}{lccccccc}
\toprule
Method &
\multicolumn{2}{c}{Benchmark Evaluation} &
\multicolumn{3}{c}{Perceptual Metrics} &
\multicolumn{2}{c}{Acceleration \& Memory} \\
\cmidrule(lr){2-3}
\cmidrule(lr){4-6}
\cmidrule(lr){7-8}

& WS Static$\uparrow$
& WS Dynamic$\uparrow$
& PSNR$\uparrow$
& SSIM$\uparrow$
& LPIPS$\downarrow$
& Speedup$\uparrow$
& Memory(GB)$\downarrow$ \\

\midrule
\multicolumn{8}{c}{\textbf{HunyuanVoyager-13B} (Huang et al., 2025) 
($512\times768$, frames = 49)} \\
\midrule

Voyager 
& 66.28 & 46.40 & $\infty$ & 1.000 & 0.000 & 1.00$\times$ & 50.44 \\

DuCa 
& 53.87 & 37.71 & 16.66 & 0.508 & 0.486 & 1.30$\times$ & 109.70 \\

ToCa 
& 47.49 & 33.24 & 15.51 & 0.409 & 0.558 & 1.01$\times$ & 107.35 \\

TaylorSeer 
& 62.46 & 43.72 & 18.32 & 0.615 & 0.293 & 0.88$\times$ & 163.79 \\

HiCache 
& 63.80 & 44.66 & 18.56 & 0.623 & 0.281 & 0.96$\times$ & 163.79 \\

TeaCache 
& 60.88 & 42.61 & 16.25 & 0.565 & 0.372 & 3.38$\times$ & 56.52 \\

EasyCache 
& 64.16 & 44.91 & 21.76 & 0.737 & 0.208 & 3.58$\times$ & 50.98 \\

HERO 
& 62.37 & 43.67 & 17.71 & 0.601 & 0.315 & 0.96$\times$ & 173.42 \\

WorldCache 
& 64.89 & 45.43 & 23.49 & 0.770 & 0.176 & 3.65$\times$ & 50.58 \\

\textbf{SpectralCache (Ours)} 
& \textbf{65.90} 
& \textbf{46.13} 
& \textbf{25.06} 
& \textbf{0.818} 
& \textbf{0.154} 
& \textbf{5.22$\times$} 
& 50.54 \\

\midrule
\multicolumn{8}{c}{\textbf{Aether-5B} (Zhu et al., 2025) 
($480\times720$, frames = 41)} \\
\midrule

Aether 
& 64.60 & 45.22 & $\infty$ & 1.000 & 0.000 & 1.00$\times$ & 46.58 \\

DuCa 
& 60.17 & 42.12 & 26.68 & 0.838 & 0.151 & 1.63$\times$ & 61.44 \\

ToCa 
& 60.15 & 42.11 & 26.68 & 0.839 & 0.151 & 1.62$\times$ & 61.78 \\

TaylorSeer 
& 57.11 & 39.97 & 22.92 & 0.713 & 0.324 & 1.66$\times$ & 77.32 \\

HiCache 
& 58.96 & 41.27 & 24.93 & 0.784 & 0.226 & 1.65$\times$ & 77.32 \\

TeaCache 
& 60.95 & 42.67 & 26.60 & 0.843 & 0.138 & 1.57$\times$ & 46.78 \\

EasyCache 
& 62.89 & 44.02 & 22.84 & 0.720 & 0.186 & 1.49$\times$ & 46.59 \\

HERO 
& 58.62 & 41.04 & 23.56 & 0.741 & 0.259 & 1.36$\times$ & 75.08 \\

WorldCache 
 & 63.68 & 44.72 & 31.87 & 0.924 & 0.066 & 1.68$\times$ & 46.59 \\

\textbf{SpectralCache (Ours)} 
& \textbf{65.44} 
& \textbf{45.82} 
& \textbf{32.73} 
& \textbf{0.959} 
& \textbf{0.050} 
& \textbf{2.20$\times$} 
& 46.59 \\

\bottomrule
\end{tabular}

\end{adjustbox}
\vspace{-0.1in}
\caption{Quantitative comparison on Image-to-World generation.
\textbf{Bold} indicates the best result.
Layer-wise methods marked with * exceed single-GPU memory limits and require CPU offloading. 
}
\vspace{-0.1in}
\label{tab:image2world}
\end{table}

\subsection{Main Results}

\textbf{Generation Results.}
% Table~\ref{tab:image2world} presents the quantitative results on HunyuanVoyager-13B and Aether-5B. 
% Our method consistently achieves the best overall generation quality with non-marginal improvements among all acceleration baselines, while providing substantial inference acceleration. 
% On HunyuanVoyager-13B, our method achieves the highest WorldScore of 65.94 and 46.16 for static and dynamic scenes, respectively, outperforming the strongest accelerated baseline, WorldCache, by 1.05 and 0.73 points. Our method also consistently improves perceptual fidelity, increasing PSNR from 23.49 to 25.06 and SSIM from 0.770 to 0.831.
% Meanwhile, it delivers a $5.22\times$ end-to-end speedup with almost unchanged GPU memory consumption (50.54GB vs.\ 50.44GB for the original model), achieving approximately $1.43\times$ higher speedup than WorldCache.
% Similar improvements are observed on Aether-5B. 
% Our method achieves the best WorldScore of 65.44 and 45.82 for static and dynamic scenes, respectively, surpassing WorldCache by 1.76 and 1.10 points. 
% It also improves perceptual fidelity, increasing PSNR from 31.87 to 32.73 and SSIM from 0.924 to 0.959. 
% Meanwhile, our method achieves a $2.20\times$ speedup, approximately 1.3x higher than the $1.68\times$ speedup of WorldCache.
% These results demonstrate that exploiting the low-rank structure and temporal evolution of singular components provides an effective alternative to conventional feature caching. 
Table~\ref{tab:image2world} presents the quantitative results on HunyuanVoyager-13B and Aether-5B.
Our method consistently achieves the best overall generation quality among all acceleration baselines while providing substantial inference acceleration.
On HunyuanVoyager-13B, our method achieves the highest WorldScores of 65.94 and 46.16 for static and dynamic scenes, respectively, outperforming WorldCache by 1.05 and 0.73 points.
It also improves perceptual fidelity, increasing PSNR from 23.49 to 25.06 and SSIM from 0.770 to 0.831, while delivering a $5.22\times$ end-to-end speedup with nearly unchanged GPU memory consumption (50.54\,GB vs.\ 50.44\,GB).
Similar improvements are observed on Aether-5B, where our method achieves the best WorldScores of 65.44 and 45.82 for static and dynamic scenes, surpassing WorldCache by 1.76 and 1.10 points.
It further improves PSNR from 31.87 to 32.73 and SSIM from 0.924 to 0.959, while achieving a $2.20\times$ speedup compared with $1.68\times$ for WorldCache.
These results demonstrate that exploiting the low-rank structure and temporal evolution of singular components effectively improves inference efficiency while preserving high generation quality.

% By reusing stable singular subspaces and efficiently approximating their evolving singular values, our method substantially reduces redundant computation while preserving the underlying feature structure and generation performance.

% \textbf{3D Reconstruction Results.}
% Table~\ref{3d_con} reports the 3D reconstruction performance on Aether-5B, including both video depth and camera pose estimation. 
% Our method achieves the best overall depth estimation performance among accelerated methods, obtaining an Abs Rel of 0.331 and accuracies of 0.520 and 0.773 under $\delta < 1.25$ and $\delta < 1.25^2$, respectively. 
% \textcolor{red}{(it seems the same as the above sentence, why mention the same results twice. maybe comment one sentence)} Compared with WorldCache, our method reduces Abs Rel from 0.341 to 0.331, while improving the two depth accuracy metrics from 0.508 and 0.741 to 0.520 and 0.773.
% For camera pose estimation, our method further reduces ATE from $0.184$ to $0.179$ and RPE trans. from 0.068 to 0.060 compared with WorldCache, while maintaining comparable RPE rot. (0.800 vs.\ 0.796). 
% More importantly, these improvements are achieved with a $3.45\times$ end-to-end speedup, approximately $1.3\times$ higher than the $2.61\times$ speedup of the strongest baseline. 
% These results demonstrate that our method effectively preserves the geometric and spatial information required for 3D world reconstruction while substantially reducing inference cost.
\begin{table}[H]
\centering
\small
\setlength{\tabcolsep}{2pt}
\renewcommand{\arraystretch}{1.15}

\begin{adjustbox}{width=\columnwidth}
\begin{tabular}{l|ccc|ccc|ccc}
\toprule
\multirow{2}{*}{\textbf{Method}}
&
\multicolumn{3}{c|}{\textbf{Video Depth Estimation}}
&
\multicolumn{3}{c|}{\textbf{Camera Pose Estimation}}
&
\multicolumn{3}{c}{\textbf{Acceleration \& Memory}}
\\
\cmidrule(lr){2-4}
\cmidrule(lr){5-7}
\cmidrule(l){8-10}
&
Abs Rel$\downarrow$
&
$\delta<1.25\uparrow$
&
$\delta<1.25^2\uparrow$
&
ATE$\downarrow$
&
RPE trans.$\downarrow$
&
RPE rot.$\downarrow$
&
Latency(s)$\downarrow$
&
Speed$\uparrow$
&
Memory
(GB)$\downarrow$
\\
\midrule
Aether      & 0.340 & 0.502 & 0.738 & 0.177 & 0.068 & 0.780 & 55.42 & 1.00$\times$ & 50.19 \\
DuCa        & 0.341 & 0.475 & 0.694 & 0.209 & 0.069 & 0.904 & 28.15 & 1.97$\times$ & 52.70 \\
ToCa        & 0.341 & 0.476 & 0.694 & 0.209 & 0.069 & 0.904 & 28.02 & 1.98$\times$ & 52.70 \\
TaylorSeer  & 0.361 & 0.460 & 0.718 & 0.197 & 0.068 & 1.134 & 26.71 & 2.07$\times$ & 58.57 \\
HiCache     & 0.346 & 0.472 & 0.712 & 0.204 & 0.069 & 1.004 & 26.51 & 2.09$\times$ & 58.57 \\
TeaCache    & 0.341 & 0.496 & 0.724 & 0.183 & 0.068 & 0.797 & 25.85 & 2.14$\times$ & 50.20 \\
EasyCache   & 0.390 & 0.479 & 0.725 & 0.183 & 0.069 & 1.061 & 27.76 & 2.00$\times$ & 50.20 \\
HERO        & 0.347 & 0.490 & 0.716 & 0.181 & 0.071 & 0.861 & 27.44 & 1.96$\times$ & 61.56 \\
Worldcache  & 0.341 & 0.508 & 0.741 & 0.184 & 0.068 & \textbf{0.796} & 21.20 & 2.61$\times$ & 50.20 \\

\textbf{SpectralCache (Ours)}
&
\textbf{0.331}
&
\textbf{0.520}
&
\textbf{0.773}
&
\textbf{0.179}
&
\textbf{0.060}
&
0.800
&
\textbf{16.06}
&
\textbf{3.45$\times$}
&
\textbf{50.20}
\\
\bottomrule
\end{tabular}
\end{adjustbox}
\vspace{-0.1in}
\caption{Quantitative comparison on 3D reconstruction on Aether.
\textbf{Bold} indicates the best result.}
\label{tab:3D_reconstruction}
\vspace{-0.15in}
\end{table}

\textbf{3D Reconstruction Results.} % \textcolor{red}{(it is too close the table 1 title, looks like table 1 title is very long including this paragraph.)}
Table~\ref{tab:3D_reconstruction} reports the 3D reconstruction performance on Aether-5B, including both video depth and camera pose estimation.
Compared with WorldCache, our method achieves the best overall depth estimation performance among accelerated methods, reducing Abs Rel from 0.341 to 0.331 while improving the accuracies under $\delta < 1.25$ and $\delta < 1.25^2$ from 0.508 and 0.741 to 0.520 and 0.773, respectively.
For camera pose estimation, our method further reduces ATE from $0.184$ to $0.179$ and RPE trans. from 0.068 to 0.060 compared with WorldCache, while maintaining comparable RPE rot. (0.800 vs.\ 0.796).
More importantly, these improvements are achieved with a $3.45\times$ end-to-end speedup, approximately $1.3\times$ higher than the $2.61\times$ speedup of the strongest baseline.
These results demonstrate that our method effectively preserves  geometric and spatial information required for 3D world reconstruction while substantially reducing inference cost.

\subsection{Ablation Studies}

\begin{table}[H]
\centering
\small

% ==================== Left Table ====================
\begin{minipage}[t]{0.48\columnwidth}
\centering
\setlength{\tabcolsep}{4pt}
\renewcommand{\arraystretch}{1.05}

\begin{tabular}{c|ccc}
\toprule
Rank
& PSNR$\uparrow$
& SSIM$\uparrow$
& Speedup$\uparrow$ \\
\midrule

Voyager
& $\infty$
& 1.000
& 1.00$\times$ \\

\midrule

Rank=32
& 24.67
& 0.806
& 5.28$\times$ \\

Rank=64
& 24.88
& 0.813
& 5.26$\times$ \\

Rank=128
& 25.06
& 0.818
& 5.22$\times$ \\

Rank=256
& 25.12
& 0.820
& 5.15$\times$ \\

\bottomrule
\end{tabular}
\caption{Ablation study on effects of ranks.}
\label{tab:rank}
\end{minipage}
\hfill
% ==================== Right Table ====================
\begin{minipage}[t]{0.48\columnwidth}
\centering
\setlength{\tabcolsep}{4pt}
\renewcommand{\arraystretch}{1.05}

\begin{tabular}{c|ccc}
\toprule
Interval
& PSNR$\uparrow$
& SSIM$\uparrow$
& Speedup$\uparrow$ \\
\midrule

Voyager
& $\infty$
& 1.000
& 1.00$\times$ \\

\midrule

$N=3$
& 25.33
& 0.822
& 4.87$\times$ \\

$N=4$
& 25.06
& 0.818
& 5.22$\times$ \\

$N=5$
& 24.72
& 0.810
& 5.44$\times$ \\

$N=6$
& 24.63
& 0.806
& 5.78$\times$ \\

\bottomrule
\end{tabular}
\caption{Ablation study on cache step intervals.}
\label{tab:interval}
\end{minipage}
% \vspace{-0.3in}
\end{table}

%\textcolor{red}{(due to space limit, maybe do not explain techniques in ablation. just discuss the results to save space. maybe not use subsubsection, use paragraph or textbf instead to save space.)}

\textbf{Effects of Different Number of Ranks.}
Table~\ref{tab:rank} studies the effect of the retained rank on generation quality and inference efficiency.
Increasing the rank generally improves generation quality at a modest cost in acceleration.
We adopt rank 128 as the default setting, which achieves a favorable trade-off with a PSNR of $25.06$, an SSIM of $0.818$, and a $5.22\times$ speedup.
Further increasing the rank to 256 yields only marginal quality improvements while slightly reducing the speedup.

% \textcolor{red}{(due to space limit, we may not have many space to discuss results. maybe just mention table or figure in ablation, and draw a conclusion. write little about the numbers from tables, or just highlight one or two numbers. no need to write long discussions with many detailed numbers. people typically just take a look at the tables and do not read the texts carefully.)}

\textbf{Effects of Different Maximum Cache Intervals.}
As shown in Table~\ref{tab:interval}, increasing the maximum cache interval $N$
improves acceleration but gradually degrades generation quality due to larger
approximation errors. We choose $N=4$ as the default setting, which achieves
a favorable trade-off between efficiency and quality, with a PSNR of $25.06$,
an SSIM of $0.818$, and a $5.22\times$ speedup. Increasing $N$ to $5$
or $6$ further improves the speedup, % to $5.44\times$ and $5.78\times$, respectively,
but leads to noticeable quality degradation.

% \begin{wraptable}[10]{r}{0.48\textwidth}
% \vspace{-0.15in}
% \centering
% \small
% \renewcommand{\arraystretch}{1.05}

% \caption{Ablation study on cache step intervals.}
% \label{tab:interval}

% \resizebox{\linewidth}{!}{
% \begin{tabular}{c|ccc}
% \toprule
% Cache interval
% & PSNR$\uparrow$ 
% & SSIM$\uparrow$ 
% & Latency$\downarrow$ \\
% \midrule

% Voyager
% & $\infty$ 
% & 1.000
% & 1.00$\times$ \\

% \midrule

% $N=3$
% & 25.33
% & 0.822
% & 4.87$\times$ \\

% $N=4$
% & 25.06
% & 0.818
% & 5.22$\times$ \\

% $N=5$
% & 24.72
% & 0.810
% & 5.44$\times$ \\

% $N=6$
% & 24.63
% & 0.806
% & 5.78$\times$ \\

% \bottomrule
% \end{tabular}
% }

% \end{wraptable}

\begin{table}[H]
% \vspace{-0.2in}
\centering
\small
\setlength{\tabcolsep}{5pt}
\renewcommand{\arraystretch}{1.05}
\begin{tabular}{c|cccccc}
\toprule
Methods
& PSNR$\uparrow$
& SSIM$\uparrow$
& WS-Static$\uparrow$
& WS-Dynamic$\uparrow$
& Latency$\downarrow$
& Memory (GB)$\downarrow$
\\
\midrule

Voyager
& $\infty$
& 1.000
& 66.28
& 46.40
& 1.00 $\times$
& 50.44
\\

HERO
& 17.71
& 0.601
& 62.37
& 43.67
& 0.96$\times$
& 173.42
\\

WorldCache
& 23.49
& 0.770
& 64.89
& 45.43
& 3.65 $\times$
& 50.58
\\

\midrule

SpectralCache (Ours)*
& \textbf{26.09}
& \textbf{0.832}
& \textbf{66.20}
& \textbf{46.31}
& 3.84 $\times$
& 50.44
\\

\textbf{SpectralCache (Ours)}
& 25.06
& 0.818
& 65.90
& 46.13
& \textbf{5.22 $\times$}
& \textbf{50.44}
\\

\bottomrule
\end{tabular}
\caption{
Ablation study on the effect of singular value scaling for full computation.
* denotes our method with singular value scaling disabled at triggered full computation steps.
% \textcolor{red}{(too close)}
%\textcolor{red}{maybe move to the same row of table 5 tp save space. If it is too wide, maybe remove memory (GB) column and use short names such as WS-D instead of WS-Dynamic, we can explain short names in table title}
}
\label{tab:without_singular}
\vspace{-0.2in}
\end{table}

\textbf{Effect of Singular Value Scaling for Full Computation.}
We further ablate the contribution of singular value scaling for full computation. As shown in Table~\ref{tab:without_singular}, 
when singular value scaling is disabled at triggered steps, 
%using only our singular value-based spectral feature caching technique 
our variant with triggered full computation achieves a even stronger  quality but with a limited speedup of $3.84\times$, compared with our $5.22\times$ speedup by enabling singular value scaling.
%achieves strong generation quality but provides a relatively limited speedup of $3.84\times$. 
%By further enabling singular value scaling to skip triggered full-computation steps, the speedup increases substantially from $3.84\times$ to $5.22\times$, while still outperforming WorldCache and HERO baselines in both perceptual and WorldScore metrics. 
These results demonstrate that singular value scaling effectively reduces expensive full backbone evaluations while maintaining high generation quality.
% \textcolor{red}{(maybe move table 3 and 4 to page 8 to distribute uniformly)}

\begin{figure}[H]
\centering

% ==================== Left: Figure ====================
\begin{minipage}[t]{0.48\columnwidth}
\vspace{-0.1in}
\centering

\includegraphics[width=\linewidth]{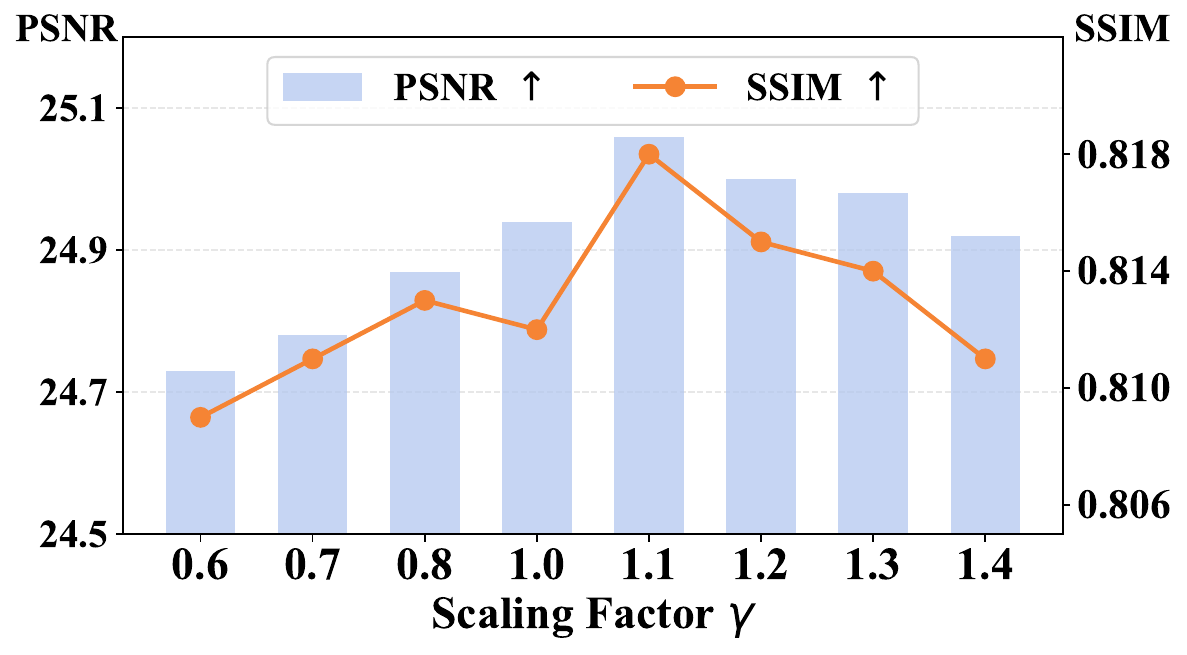}
\vspace{-0.3in}
\captionof{figure}{Ablation study on scaling factors $\gamma$.}
\label{fig:gamma}
\end{minipage}
\hfill
% ==================== Right: Table ====================
\begin{minipage}[t]{0.51\columnwidth}
\vspace{0pt}
\centering
\scriptsize
\setlength{\tabcolsep}{3pt}
\renewcommand{\arraystretch}{0.9}

\resizebox{\linewidth}{!}{
\begin{tabular}{l|c}
\toprule
Operation & Avg. Latency Overhead (ms) \\
\midrule
Full DiT Computation & 17942.06 \\
\midrule
SVD Decomposition & 82.16\\
Extrapolation Cache Step & 16.40 \\
Scaling Approximation Step & 45.57 \\
Error Computation & 0.33\\
\bottomrule
\end{tabular}
}

\captionof{table}{Average latency overhead of different operations in SpectralCache pipeline for HunyuanVoyager-13B.}
\label{tab:cost_analysis}

\end{minipage}
\vspace{-0.2in}
\end{figure}

\textbf{Effects of Different Singular Value Scaling Factors.}
As discussed in Section~\ref{motivation}, cache-triggered full-computation steps exhibit stable singular subspaces and an approximately constant singular-value scaling factor of $1.1$ relative to the preceding full-computation step.
We therefore ablate the scaling factor $\gamma$ in Figure~\ref{fig:gamma}.
The best performance is achieved at $\gamma=1.1$, with a PSNR of $25.06$ and an SSIM of $0.818$, consistent with our empirical observation.
%Performance remains relatively stable around this value but gradually degrades as $\gamma$ deviates further from $1.1$.
%Therefore, we use $\gamma=1.1$ as the default setting in all experiments.

% \begin{wrapfigure}[11]{r}{0.5\textwidth}
% \vspace{-0.2in}
% \centering
% \includegraphics[width=\linewidth]{Figure/gamma_ablation.pdf}
% \vspace{-0.3in}
% \caption{
% % Singular-subspace consistency across full computations.
% Ablation study on scaling factors $\gamma$.
% }
% \label{fig:gamma}
% \vspace{-0.2in}
% \end{wrapfigure}

% \begin{wraptable}[9]{r}{0.48\textwidth}
% \vspace{-0.2in}
% \centering
% \small
% \setlength{\tabcolsep}{10pt}
% \renewcommand{\arraystretch}{1.05}
% \caption{Time overhead of different operations in our pipeline for HunyuanVoyager-13B.}
% \label{tab:cost_analysis}
% \vspace{-0.1in}
% \resizebox{\linewidth}{!}{
% \begin{tabular}{l|c}
% \toprule
% Operation & Speedup $\uparrow$ (rel. to Full) \\
% \midrule
% Full DiT Computation & 1.0$\times$ \\
% \midrule
% SVD Decomposition & 221.9$\times$ \\
% Extrapolation Cache Step & 1112.3$\times$ \\
% Scaling Approximation Step & 400.0$\times$ \\
% Error Computation & 55527.3$\times$ \\
% \bottomrule
% \end{tabular}
% }

% \end{wraptable}

\subsection{Cost Analysis}
\label{sec:cost_analysis}

We further analyze the additional computational and memory overhead introduced by our spectral caching pipeline.
%Throughout the entire generation process of HunyuanVoyager-13B, 
Our method only needs to retain the low-rank SVD states from  two most recent full computation steps,
resulting in approximately $0.1$\,GB additional cache memory, which accounts for only about $0.2\%$ of the full model with 50.44 GB.
%the 50.44GB overhead of the full model.
% In terms of computational overhead, each SVD decomposition takes approximately
% 82.2ms, while an extrapolation-based cache step and a scaling-based
% approximation step require approximately 16.4ms and 45.6ms,
% respectively.
% The approximation error computation introduces only about 0.33ms of additional overhead.
We compare the computational costs of the SVD and approximation operations with that of a single full DiT computation, as shown in Table~\ref{tab:cost_analysis}.
The additional memory and computational costs introduced by SVD,
spectral approximation, and error estimation are minor compared with the
expensive full DiT computation, while enabling substantial reductions in the
number of full backbone evaluations.

\section{Conclusion}

% \textcolor{red}{@Zhendong}

In this paper, we present \textbf{SpectralCache}, a training-free spectral feature caching framework for accelerating diffusion-based world models.
Unlike existing caching methods that primarily exploit temporal redundancy at the feature or token level, SpectralCache leverages the underlying spectral structure of diffusion features.
We reveal that world-model features exhibit stable singular subspaces across nearby denoising steps, while their singular values follow predictable evolution patterns.
Based on these observations, SpectralCache reuses stable singular subspaces and efficiently estimates singular values through linear extrapolation for consecutive cache steps and singular value scaling for cache-triggered full computation steps, substantially reducing expensive backbone evaluations.
Extensive experiments on representative world models demonstrate that SpectralCache consistently improves both generation quality and inference efficiency over existing training-free caching methods, achieving up to $5.22\times$ acceleration while preserving high generation quality.

\section*{AI use statement}
In this work, large language models (LLMs) were used solely as a general-purpose writing assistant. Their role was limited to correcting grammar, fixing typographical errors, and polishing the language for clarity and readability.
% Overall, SpectralCache provides a simple and effective spectral perspective for accelerating diffusion-based world models.

% \begin{table}[t]
% \centering
% \small
% \setlength{\tabcolsep}{10pt}
% \renewcommand{\arraystretch}{1.05}

% \caption{Ablation study on scaling factors $\gamma$.\textcolor{red}{(switch to a figure may be better for presentation)}}
% \label{tab:gamma}

% \begin{tabular}{c|cc}
% \toprule
% Magnitude of $\gamma$
% & PSNR$\uparrow$ 
% & SSIM$\uparrow$ 
% \\
% \midrule

% Voyager
% & $\infty$ 
% & 1.000
%  \\

% \midrule

% $\gamma$=0.6
% & 24.73
% &  0.809
% \\

% $\gamma$=0.7
% & 24.78
% &  0.811
% \\

% $\gamma$=0.8
% & 24.80
% &  0.813
% \\

% $\gamma$=1.0
% & 24.94
% &  0.812
% \\

% $\gamma$=1.1
% & \textbf{25.06}
% & \textbf{0.818}
% \\

% $\gamma$=1.2
% & 25.00
% &  0.815
% \\

% $\gamma$=1.3
% & 24.98
% &  0.814
% \\

% $\gamma$=1.4
% & 24.92
% &  0.811
% \\

% \bottomrule
% \end{tabular}

% \end{table}

\bibliography{iclr2027_conference}
\bibliographystyle{iclr2027_conference}

\newpage

\appendix
\section{Appendix}

\begin{figure}[H]
\vspace{-0.1in}
\centering
\includegraphics[width=0.9\linewidth]{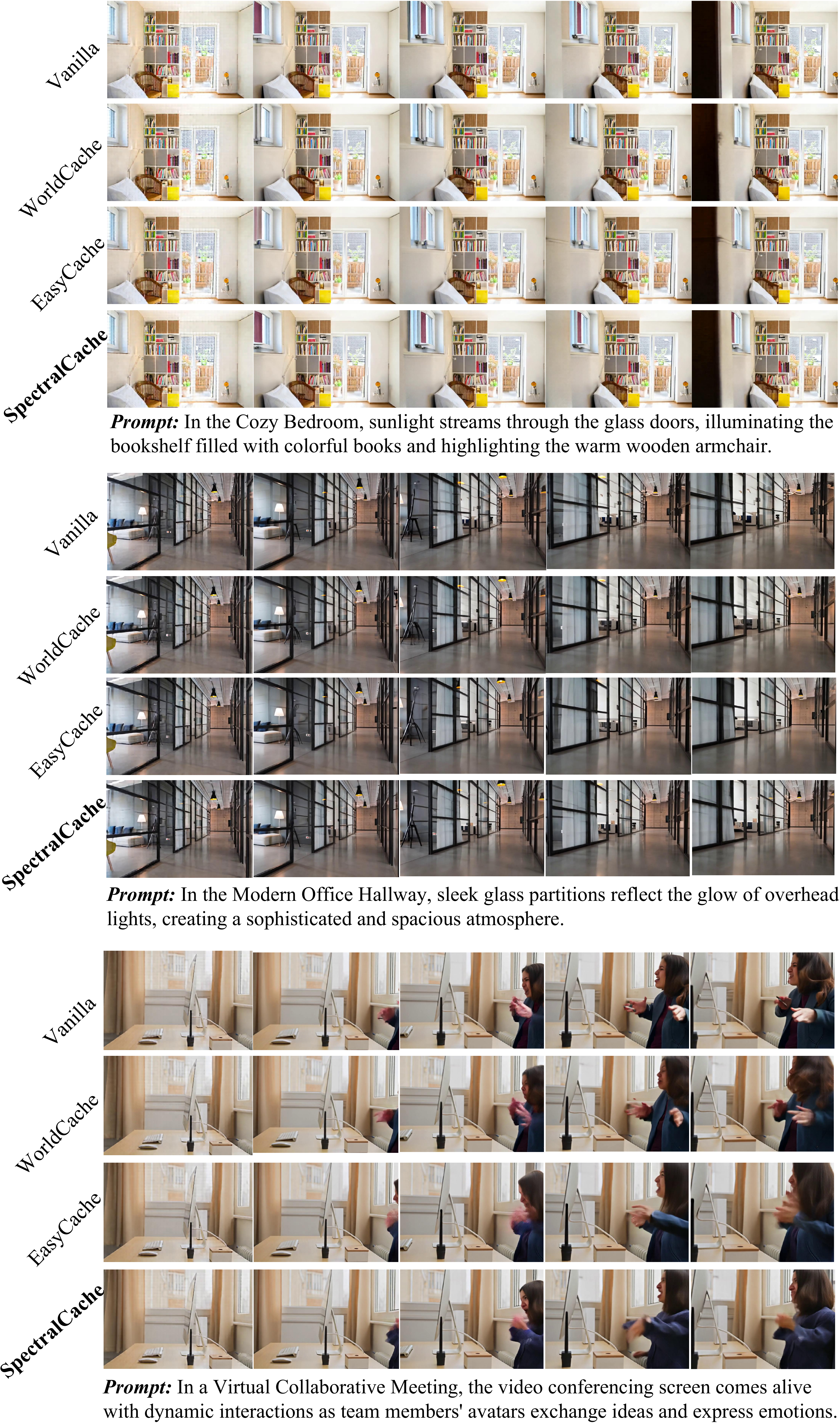}
\vspace{-0.15in}
    \caption{
More visual comparison on HunyuanVoyager-13B between SpectralCache and existing methods.
}
\label{generation_hunyuan}
\vspace{-0.2in}
\end{figure}

\subsection{More Visualization Results.}

\begin{figure}[H]
\vspace{-0.1in}
\centering
\includegraphics[width=0.98\linewidth]{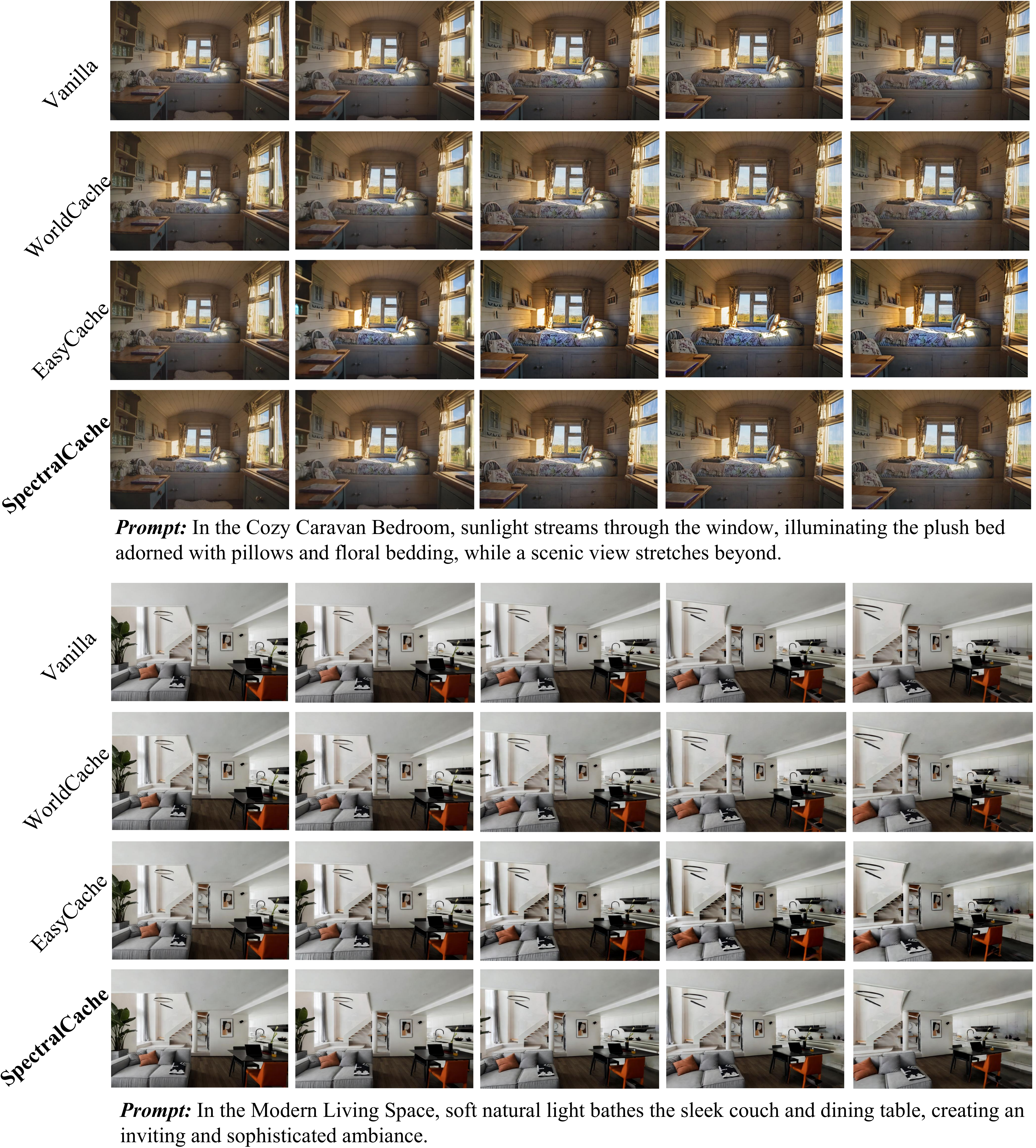}
%\vspace{-0.25in}
    \caption{
More visual comparison on Aether-5B between SpectralCache and existing methods.
}
\label{generation_aether}
\vspace{-0.2in}
\end{figure}

Here, we provide more visual comparisons to demonstrate the effectiveness of our proposed SpectralCache. The results are
shown in Figure \ref{generation_hunyuan} and Figure \ref{generation_aether}.

\subsection{Overall Framework}

We consider a diffusion-based world model that generates video and geometric
modalities through a sequence of denoising steps.
Let $\mathbf{z}_t$ denote the noisy latent at step $t$. %where the steps are ordered from high to low noise. 
Conditioned on text, camera trajectories,
and other control signals $\mathbf{c}$, a diffusion Transformer evaluates
\begin{equation}
    \mathbf{F}_t
    = \mathcal{G}_{\theta}(\mathbf{z}_t, t, \mathbf{c}),
    \qquad
    \mathbf{y}_t
    = \mathcal{H}_{\theta}(\mathbf{F}_t),
    \label{eq:dit_feature_and_head}
\end{equation}
where $\mathbf{F}_t$ denotes the final-layer visual features, % produced by the Transformer backbone, 
$\mathcal{G}_{\theta}$ represents the expensive
Transformer computation, and $\mathcal{H}_{\theta}$ projects the feature into %the corresponding denoising prediction 
denoised $\mathbf{y}_t$.
The scheduler then updates the latent as %\textcolor{red}{(the step is decreasing here. But in following, the step number is increasing, maybe use consistent)}
\begin{equation}
    \mathbf{z}_{t+1}
    = \mathcal{S}(\mathbf{z}_t, \mathbf{y}_t, t).
    \label{eq:diffusion_scheduler_update}
\end{equation}
%Since the Transformer backbone is repeatedly computed throughout the denoising process, 
$\mathcal{G}_{\theta}$ dominates the overall inference time cost due to repeated denoising precess. Following the description and notations in main paper, the overall framework of our algorithm can be shown as:
%Existing caching methods reduce this cost by replacing selected backbone evaluations with inexpensive feature cache approximations. We refer to a step that performs the complete backbone evaluation as a \emph{full-computation step}. 

\label{app:algorithm}

\begin{algorithm}[H]
\caption{Spectral Feature Caching for Diffusion-Based World Models 
%\textcolor{red}{(maybe move to appendix, can only keep a figure or algorithm due to space)}
}
\label{alg:spectral_reuse}
\begin{algorithmic}[1]

\REQUIRE Initial latent $\mathbf{z}_T$, backbone $\mathcal{G}_{\theta}$,
output head $\mathcal{H}_{\theta}$, scheduler $\mathcal{S}$,
rank $r$, error threshold $\tau$, maximum cache interval $N$,
scaling factor $\gamma$

\STATE Initialize spectral buffer $\mathcal{B}\leftarrow\varnothing$
\STATE $E_{\mathrm{acc}}\leftarrow0$, $k\leftarrow0$
\STATE $\mathrm{last\_refresh\_true}\leftarrow\mathrm{False}$

\FOR{$t=T,T-1,\ldots,1$}    %check

    \IF{$|\mathcal{B}|<2$}
        \STATE $\mathbf{X}_t\leftarrow
        \mathcal{G}_{\theta}(\mathbf{z}_t,t,\mathbf{c})$,
        Compute
        $(\mathbf{U}_t^r,\mathbf{\Sigma}_t^r,
        \mathbf{V}_t^r,\mathbf{R}_t)$
        \STATE $\mathcal{Q}_t\leftarrow
        \{\mathbf{U}_t^r,\mathbf{\Sigma}_t^r,
        \mathbf{V}_t^r,\mathbf{R}_t,t\}$
        \STATE Append $\mathcal{Q}_t$ to $\mathcal{B}$ and keep the latest two states
        \STATE $E_{\mathrm{acc}}\leftarrow0$, $k\leftarrow0$
        \STATE $\mathrm{last\_refresh\_true}\leftarrow\mathrm{True}$

    \ELSE
        \STATE Let $\mathcal{Q}_m,\mathcal{Q}_n$ be the two latest
        true full-computation states in $\mathcal{B}$

        \IF{$E_{\mathrm{acc}}<\tau$ \AND $k<N$}

            \STATE $\beta_t\leftarrow\dfrac{t-n}{n-m}$,
         $\widehat{\mathbf{\Sigma}}_t^r
            \leftarrow
            \mathbf{\Sigma}_n^r
            +
            \beta_t
            \left(
            \mathbf{\Sigma}_n^r-\mathbf{\Sigma}_m^r
            \right)$,
             $\widehat{\mathbf{X}}_t
            \leftarrow
            \mathbf{U}_n^r
            \widehat{\mathbf{\Sigma}}_t^r
            (\mathbf{V}_n^r)^\top
            +
            \mathbf{R}_n$

            \STATE Compute chaotic-token error $E(t)$:
            \STATE $E_{\mathrm{acc}}\leftarrow E_{\mathrm{acc}}+E(t)$,
            $k\leftarrow k+1$,
             $\mathbf{X}_t\leftarrow\widehat{\mathbf{X}}_t$

        \ELSE
            \IF{$\mathrm{last\_refresh\_true}=\mathrm{True}$}
                \STATE $\widehat{\mathbf{\Sigma}}_t^r
                \leftarrow
                \gamma\mathbf{\Sigma}_n^r$,
                 $\widehat{\mathbf{X}}_t
                \leftarrow
                \mathbf{U}_n^r
                \widehat{\mathbf{\Sigma}}_t^r
                (\mathbf{V}_n^r)^\top
                +
                \mathbf{R}_n$,
                 $\mathbf{X}_t\leftarrow\widehat{\mathbf{X}}_t$
                 \STATE $E_{\mathrm{acc}}\leftarrow0$, $k\leftarrow0$
                \STATE $\mathrm{last\_refresh\_true}\leftarrow\mathrm{False}$

            \ELSE
                \STATE $\mathbf{X}_t\leftarrow
                \mathcal{G}_{\theta}(\mathbf{z}_t,t,\mathbf{c})$,
             Compute
                $(\mathbf{U}_t^r,\mathbf{\Sigma}_t^r,
                \mathbf{V}_t^r,\mathbf{R}_t)$

                \STATE $\mathcal{Q}_t\leftarrow
                \{\mathbf{U}_t^r,\mathbf{\Sigma}_t^r,
                \mathbf{V}_t^r,\mathbf{R}_t,t\}$

                \STATE Append $\mathcal{Q}_t$ to $\mathcal{B}$ and keep the latest two states
                \STATE $E_{\mathrm{acc}}\leftarrow0$, $k\leftarrow0$
                \STATE $\mathrm{last\_refresh\_true}\leftarrow\mathrm{True}$
            \ENDIF
        \ENDIF
    \ENDIF

    \STATE $\mathbf{y}_t\leftarrow
    \mathcal{H}_{\theta}(\mathbf{X}_t)$
    \STATE $\mathbf{z}_{t-1}\leftarrow
    \mathcal{S}(\mathbf{z}_t,\mathbf{y}_t,t)$

\ENDFOR
\end{algorithmic}
\end{algorithm}

Algorithm~\ref{alg:spectral_reuse} summarizes the complete inference pipeline of SpectralCache.
Given the initial latent, diffusion backbone, output head, and scheduler, we first perform two full backbone computations to initialize the spectral buffer $\mathcal{B}$.
For each full computation feature, we apply low-rank decomposition and store its singular subspaces, singular values, and residual, while retaining only the two most recent spectral states.
These states provide the spectral information required for subsequent feature approximation without repeatedly evaluating the expensive diffusion backbone.
After initialization, SpectralCache dynamically switches among spectral extrapolation, singular-value scaling, and true full computation according to the accumulated approximation error and cache interval.
When the accumulated error remains below the threshold $\tau$ and the number of consecutive cache steps $k$ is smaller than the maximum interval $N$, the current feature is approximated through spectral extrapolation.
Specifically, we reuse the stable singular subspaces from the latest full computation state and linearly extrapolate its singular values from the two most recent full computation states.
The corresponding residual is then added to reconstruct the current feature.
After each approximation, the error of chaotic tokens is evaluated and accumulated to determine whether spectral extrapolation can continue.
Once the accumulated error exceeds $\tau$ or the cache interval reaches $N$, a refresh is triggered to prevent excessive approximation error.
Rather than directly performing an expensive backbone evaluation at every triggered step, SpectralCache further exploits the spectral consistency between neighboring full computation features.
If the preceding refresh is a true full computation, the current triggered step is approximated by reusing its singular subspaces and residual while scaling the singular values with factor $\gamma$, thereby skipping another expensive backbone evaluation.
Otherwise, a true full computation is performed, and its newly decomposed spectral state is inserted into $\mathcal{B}$ to refresh the extrapolation basis.
The approximation error and cache counter are then reset.
Finally, the resulting feature is passed through the output head and diffusion scheduler to obtain the latent for the next denoising step.

\subsection{More details for Experiments}
\label{experiments}

\textbf{Detailed Experimental Settings.}
Following WorldCache~\cite{feng2026worldcache}, we evaluate on two
multi-modal diffusion world models, HunyuanVoyager-13B and Aether-5B.
For HunyuanVoyager-13B, we use the standard 50-step inference protocol
to generate 49-frame videos at $512\times768$ resolution.
For Aether-5B, we generate 41-frame videos at $480\times720$ resolution
with 50 denoising steps, while using 30 steps for 3D reconstruction.
Following the default configuration of \cite{feng2026worldcache}, we set the token-grouping
percentiles to static token $p_s=0.3$ and chaotic token $p_c=0.7$, and use an accumulated-error
threshold of $\tau=1.0$ for chaotic tokens.
We evaluate generation quality using WorldScore and perceptual consistency
against the no-cache baseline using PSNR and SSIM.
All methods use identical conditioning inputs and sampling schedules for
fair comparison.
For WorldScore benchmark~\cite{duan2025worldscore}, we follow its standard evaluation
protocol, which formulates world generation as sequential next-scene
generation under explicit scene and camera-trajectory specifications.
WorldScore evaluates three aspects: controllability, generation quality,
and dynamics. We report WorldScore-Static, which aggregates controllability
and quality metrics, and WorldScore-Dynamic, which further incorporates
motion accuracy, magnitude, and smoothness. For generation results evaluation, we follow the standard benchmark protocol \cite{feng2026worldcache,duan2025worldscore,song2025hero} across
diverse indoor/outdoor scenarios and realistic/stylized conditions.
We report both WorldScore-Static and WorldScore-Dynamic.

\textbf{Effect of Cache Error Threshold.}
Table~\ref{tab:error_ablation} studies the effect of the accumulated chaotic-token error threshold $\tau$, which controls when the accumulated approximation error becomes sufficiently large to terminate the current cache sequence and trigger a refresh computation.
In our main experiments, we directly adopt $\tau=1.0$, the default setting of WorldCache, without additional tuning for SpectralCache.
A smaller $\tau$ imposes a stricter error constraint and therefore triggers full computations more frequently, resulting in higher generation quality but lower acceleration.
For example, $\tau=0.5$ achieves the highest PSNR of 25.39, but reduces the speedup to $4.42\times$.
As $\tau$ increases, SpectralCache tolerates larger accumulated errors and performs more aggressive approximation, increasing the speedup from $4.83\times$ at $\tau=0.7$ to $5.36\times$ at $\tau=1.5$.
However, overly large thresholds gradually degrade generation quality, with PSNR decreasing to 24.71 at $\tau=1.5$.
The inherited default $\tau=1.0$ achieves a PSNR of 25.06 with a $5.22\times$ speedup, providing a favorable balance between generation fidelity and inference efficiency.
These results also demonstrate that SpectralCache performs robustly across a reasonable range of error thresholds without requiring method-specific tuning of $\tau$.

\begin{table}[H]
\centering
\small
\setlength{\tabcolsep}{12pt}
\renewcommand{\arraystretch}{1.15}

\caption{Ablation study on different cache error threshold $\tau$ .}
\label{tab:error_ablation}

\begin{tabular}{c|cc}
\toprule
$\tau$ 
& PSNR$\uparrow$ 
& Latency (rel. to base)$\downarrow$ \\
\midrule

EasyCache
& 21.76
& 3.58 $\times$ \\
\midrule

0.5
& 25.39
& 4.42$\times$ \\

0.7
& \textbf{25.06}
&  4.83$\times$\\

1.0
& 25.06
& 5.22$\times$ \\

1.3
& 24.80
& 5.32$\times$ \\

1.5
& 24.71
& 5.36$\times$ \\

\bottomrule
\end{tabular}

\end{table}

\textbf{Token Grouping Percentiles.}
We further study the sensitivity of SpectralCache to the token grouping percentiles $p_s$ and $p_c$ adopted from~\cite{feng2026worldcache}, which control the partition of tokens into static, regular, and chaotic groups according to their temporal variations.
Specifically, $p_s$ determines the fraction of tokens regarded as static, while $p_c$ specifies the percentile threshold above which tokens are treated as chaotic.
For example, $\{p_s,p_c\}=\{0.3,0.8\}$ assigns the lowest 30\% of tokens to the static group and the highest 20\% to the chaotic group, with the remaining tokens treated as regular.
As shown in Table~\ref{tab:token_grouping}, SpectralCache consistently achieves strong perceptual quality across different grouping configurations, indicating that our spectral caching framework is relatively robust to the choice of token grouping percentiles.
Among them, $\{p_s,p_c\}=\{0.3,0.7\}$ achieves the best overall performance, reaching a PSNR of $25.06$ and an SSIM of $0.818$.
Changing either threshold leads to a modest quality degradation; for instance, increasing $p_c$ from $0.7$ to $0.8$ decreases PSNR from $25.06$ to $24.84$, while reducing $p_s$ from $0.3$ to $0.2$ results in lower PSNR and SSIM.
Therefore, consistent with the default grouping strategy in~\cite{feng2026worldcache}, we adopt $p_s=0.3$ and $p_c=0.7$ for SpectralCache in our main experiments.

\begin{table}[H]
\centering
\small
\setlength{\tabcolsep}{12pt}
\renewcommand{\arraystretch}{1.15}

\caption{Ablation study on token grouping percentiles.}
\label{tab:token_grouping}

\begin{tabular}{c|cc}
\toprule
$\{p_s, p_c\}$ 
& PSNR$\uparrow$ 
& SSIM$\uparrow$ \\
\midrule

EasyCache
& 21.76
& 0.737 \\
\midrule

$\{0.3, 0.8\}$
& 24.84
& 0.813 \\

$\{0.3, 0.7\}$
& \textbf{25.06}
& \textbf{0.818} \\

$\{0.3, 0.6\}$
& 24.98
& 0.814 \\

$\{0.2, 0.8\}$
& 24.81
& 0.808 \\

$\{0.2, 0.7\}$
& 24.94
& 0.811 \\

\bottomrule
\end{tabular}

\end{table}

\end{document}